\documentclass[10pt,twocolumn,letterpaper]{article}

\usepackage{cvpr}              

\definecolor{cvprblue}{rgb}{0.21,0.49,0.74}
\usepackage[pagebackref,breaklinks,colorlinks,allcolors=cvprblue]{hyperref}
\usepackage[accsupp]{axessibility} 
\usepackage{multirow}
\usepackage{pifont}
\usepackage{colortbl}
\usepackage{subcaption}
\usepackage{tabularx}
\def\paperID{5427} 
\def\confName{CVPR}
\def\confYear{2025}

\title{Object-Shot Enhanced Grounding Network for Egocentric Video 
}

\author{
Yisen Feng$^{1}$\quad Haoyu Zhang$^{1\,2}$\quad Meng Liu$^{3}\thanks{represents corresponding author.}$\quad Weili Guan$^{1}$\quad Liqiang Nie$^{1*}$\\
$^1$Harbin Institute of Technology (Shenzhen) \quad  $^2$Pengcheng Laboratory    \\$^3$Shandong Jianzhu University\\
{\tt\small \{yisenfeng.hit, zhang.hy.2019, mengliu.sdu, honeyguan, nieliqiang\}@gmail.com} 
}

\begin{document}
\maketitle
\begin{abstract}
Egocentric video grounding is a crucial task for embodied intelligence applications, distinct from exocentric video moment localization. Existing methods primarily focus on the distributional differences between egocentric and exocentric videos but often neglect key characteristics of egocentric videos and the fine-grained information emphasized by question-type queries. To address these limitations, we propose OSGNet, an Object-Shot enhanced Grounding Network for egocentric video. Specifically, we extract object information from videos to enrich video representation, particularly for objects highlighted in the textual query but not directly captured in the video features. Additionally, we analyze the frequent shot movements inherent to egocentric videos, leveraging these features to extract the wearer's attention information, which enhances the model's ability to perform modality alignment. Experiments conducted on three datasets demonstrate that OSGNet achieves state-of-the-art performance, validating the effectiveness of our approach.
Our code can be found at \url{https://github.com/Yisen-Feng/OSGNet}.

\end{abstract}    
\section{Introduction}
\label{sec:intro}
\begin{figure}
    \centering
    \begin{subfigure}{\linewidth}
        \centering
        \includegraphics[width=\textwidth]{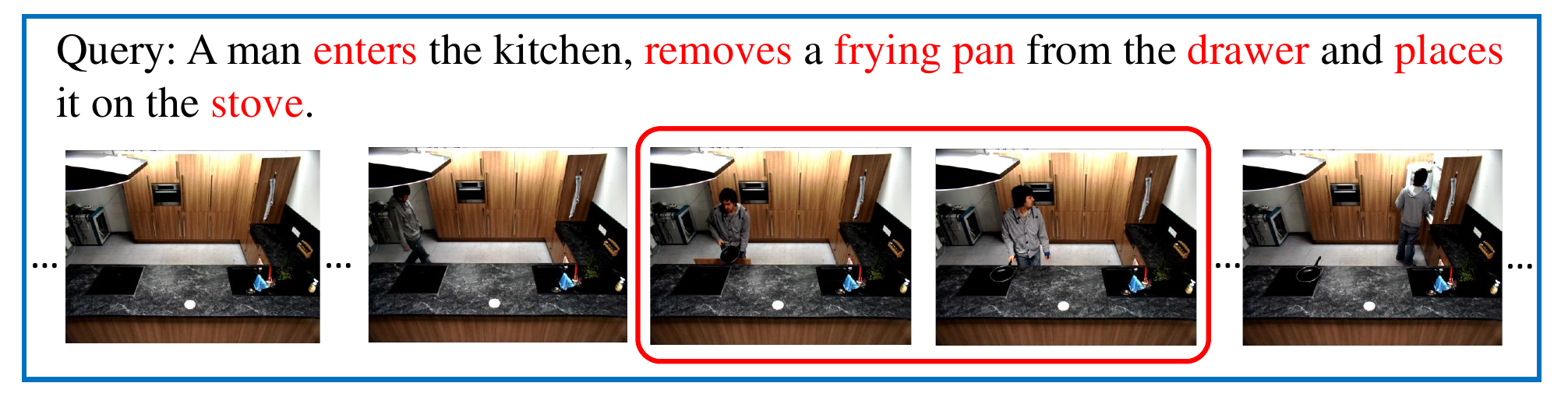} 
        \caption{Exocentric video moment localization task from TACoS~\cite{regneri2013grounding}}
        \label{fig:sub1}
    \end{subfigure}
    
    \begin{subfigure}{\linewidth}
        \centering
        \includegraphics[width=\textwidth]{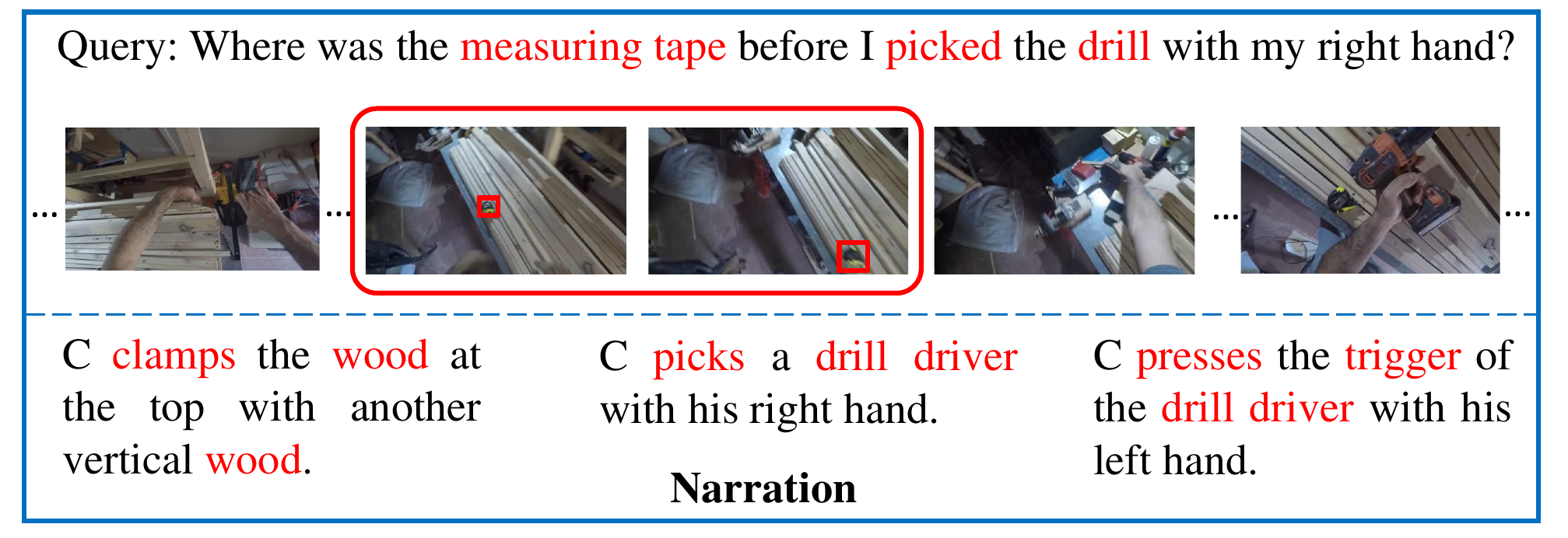} 
        \caption{Natural language query task and narrations from Ego4D~\cite{grauman2022ego4d}}
        \label{fig:sub2}
    \end{subfigure}
    \vspace{-4ex}
   \caption{Illustration of exocentric and egocentric video grounding, accompanied by annotated narrations for the egocentric video. Key verbs and nouns are highlighted in red.}
    \vspace{-2ex}
    \label{fig: example}
\end{figure}

With advancements in wearable camera technology,  Ego4D~\cite{grauman2022ego4d} introduces the Natural Language Query (NLQ) task for egocentric video grounding. NLQ aims to identify the specific video moment that answers a question-type query within an untrimmed egocentric video, as shown in Figure~\ref {fig: example}(b). The dynamic and complex camera perspectives in egocentric videos~\cite{pmlr-v235-zhang24aj} make video comprehension~\cite{li2023fine,li2023mask} significantly more challenging compared to the fixed viewpoints of exocentric video grounding, as shown in Figure~\ref{fig: example}(a). Moreover, NLQ queries often focus on fine-grained details of background objects (e.g., ``measuring tape'' in Figure~\ref{fig: example}(b)), unlike exocentric video grounding tasks that emphasize character actions (e.g., ``enters'' and ``removes'' in Figure~\ref{fig: example}(a)), adding complexity to mining background information from video. Consequently, despite progress in exocentric video grounding, existing methods struggle with the unique challenges of NLQ. Nonetheless, NLQ enables innovative applications~\cite{10.1145/3474085.3475234,10239469}, such as smart assistant systems and memory retrieval modules for autonomous robots, underscoring the urgent need to address these challenges.

Though existing methods~\cite{chen2022internvideo,lin2022egocentric,pramanick2023egovlpv2,pei2024egovideo,ramakrishnan2023naq} for NLQ have made significant advances in video-text pretraining, \textbf{they fail to address the issue of video features lacking the fine-grained object information needed to answer detailed queries}. 
As shown in Figure~\ref{fig: example}(b), egocentric video-text pretraining datasets generally provide narrations focusing on character actions involving objects. These datasets, combined with clip-narration contrastive learning, lead current video backbones to overlook fine-grained background object details in the extracted features. In contrast, NLQ emphasizes enhancing people's memory experience, often involving queries centered on background objects that are not part of active interactions, such as the ``measuring tape'' placed casually on the table, as illustrated in Figure~\ref{fig: example}(b). 

 
Moreover, existing methods \textbf{fail to fully leverage the rich attention information embedded in egocentric videos.} 
 Egocentric videos feature frequent camera movements, as the camera is typically worn on the head, moving with the wearer’s actions. This movement implicitly encodes head motion information, signaling shifts in the wearer’s attention and focus, an aspect that is crucial for NLQ but often overlooked by current methods. As shown in Figure~\ref{fig: example}(b), 
  initially, the individual fixes the wood, then walks to pick up a drill driver, and finally uses the drill driver to drill holes in the wood.
 These behaviors are independent yet interconnected. Capturing these attention shift points is vital for clarifying video structure and improving video understanding.

Building upon these findings, we propose a novel \textbf{O}bject-\textbf{S}hot enhanced \textbf{G}rounding \textbf{Net}work (OSGNet) for egocentric videos, addressing the unique challenges posed by the NLQ task. \textbf{To address the challenge of insufficient fine-grained object information}, we introduce an object extraction process (see Figure~\ref{fig: model}(a)), which captures detailed object-level information. These features are then integrated using a multi-modal fusion mechanism that employs parallel cross-attention within the main branch. This strategy ensures that both visual and textual cues are effectively leveraged to enhance localization accuracy. 
\textbf{To exploit the dynamics of the wearer's attention}, we extract head-turning data from egocentric video, which provides insights into shifts in the wearer's focus. Based on these shifts, we segment the video into semantically distinct shots and apply contrastive learning to strengthen the model's ability to align these shots with the query. This approach enhances the model's capacity to capture and utilize attention-driven context for improved video grounding. 
We evaluate our model on three benchmark datasets: Ego4D-NLQ, Ego4D-Goal-Step~\cite{song2024ego4d}, and TACoS~\cite{regneri2013grounding}. On Ego4D-NLQ, OSGNet outperforms GroundVQA~\cite{di2024grounded}, achieving a 2.15\% improvement in Rank@1 at IoU=0.5. On Ego4D-Goal-Step, OSGNet surpasses BayesianVSLNet~\cite{plou2024carlor} with a 3.65\% increase in Rank@1 at IoU=0.3. On TACoS, OSGNet achieves a 3.32\% improvement in Rank@1 at IoU=0.5 over SnAG~\cite{mu2024snag}. These results underscore the effectiveness of our model in enhancing video grounding.

\noindent \textbf{Contributions.} 1) We integrate fine-grained object information into the egocentric video grounding task, thereby improving the accuracy of background object-related query localization. 2) We introduce a wearer movement-aware shot branch that leverages shot-level contrastive learning, collaborating with the main branch to further enhance egocentric video grounding performance.
And 3) our OSGNet outperforms current state-of-the-art approaches across three benchmark datasets, setting a new standard for egocentric video grounding.



\section{Related Work}
\label{sec:Related Work}

\subsection{Video Moment Localization}
Video moment localization aims to determine the start and end timestamps of a video moment in response to a natural language query. Existing approaches~\cite{xiao2024bridging,qu2024chatvtg,tan2024siamese,zhang2023helping,liu2018attentive,liu2018cross} can be broadly classified into proposal-based and proposal-free methods, which primarily differ in their strategies for generating candidate moments. Proposal-based methods generally follow a two-stage process involving the generation of candidate moment features and subsequently matching them with textual queries. Sliding windows have been extensively employed for generating candidates \cite{Hendricks2017, hu2021coarse, Hu2021}, with approaches like \cite{Gao2017a} incorporating regression to refine window boundaries. Subsequent work \cite{xu2019multilevel, chen2019semantic, Xiao2021} enhances candidate quality by integrating text information at an early stage, while others \cite{zhang2020b, zhang2021multi} increased the density of candidate moments through exhaustive enumeration.

On the other hand, proposal-free methods aim to directly generate features for individual moments or the entire video, leading to fewer features and increased computational efficiency. Some approaches \cite{li2022compositional, zhou2021embracing, mun2020local} extract features for the entire video sequence, whereas others \cite{zhang2020, fang2023you} focus on predicting the probability of a particular moment being part of the desired query segment. The emergence of new datasets involving long video sequences \cite{grauman2022ego4d, soldan2022mad} has posed significant challenges for existing methods, which often struggle with performance degradation on longer videos. To address these challenges, \cite{hou2023cone} proposes a query-guided window selection strategy under the assumption that short windows are sufficient, while \cite{mu2024snag} presents a single-stage approach that aligns multiple video moments without making similar assumptions.

As discussed in Section~\ref{sec:intro}, significant differences in the distribution of videos and queries between the exocentric video moment localization task and the egocentric video NLQ task hinder the direct applicability of existing frameworks. To overcome these challenges, our model incorporates fine-grained object-level information and employs shot-level contrastive learning, thereby enhancing the accuracy of egocentric video grounding.

\begin{figure*}[t]
  \centering
  \includegraphics[width=\linewidth]{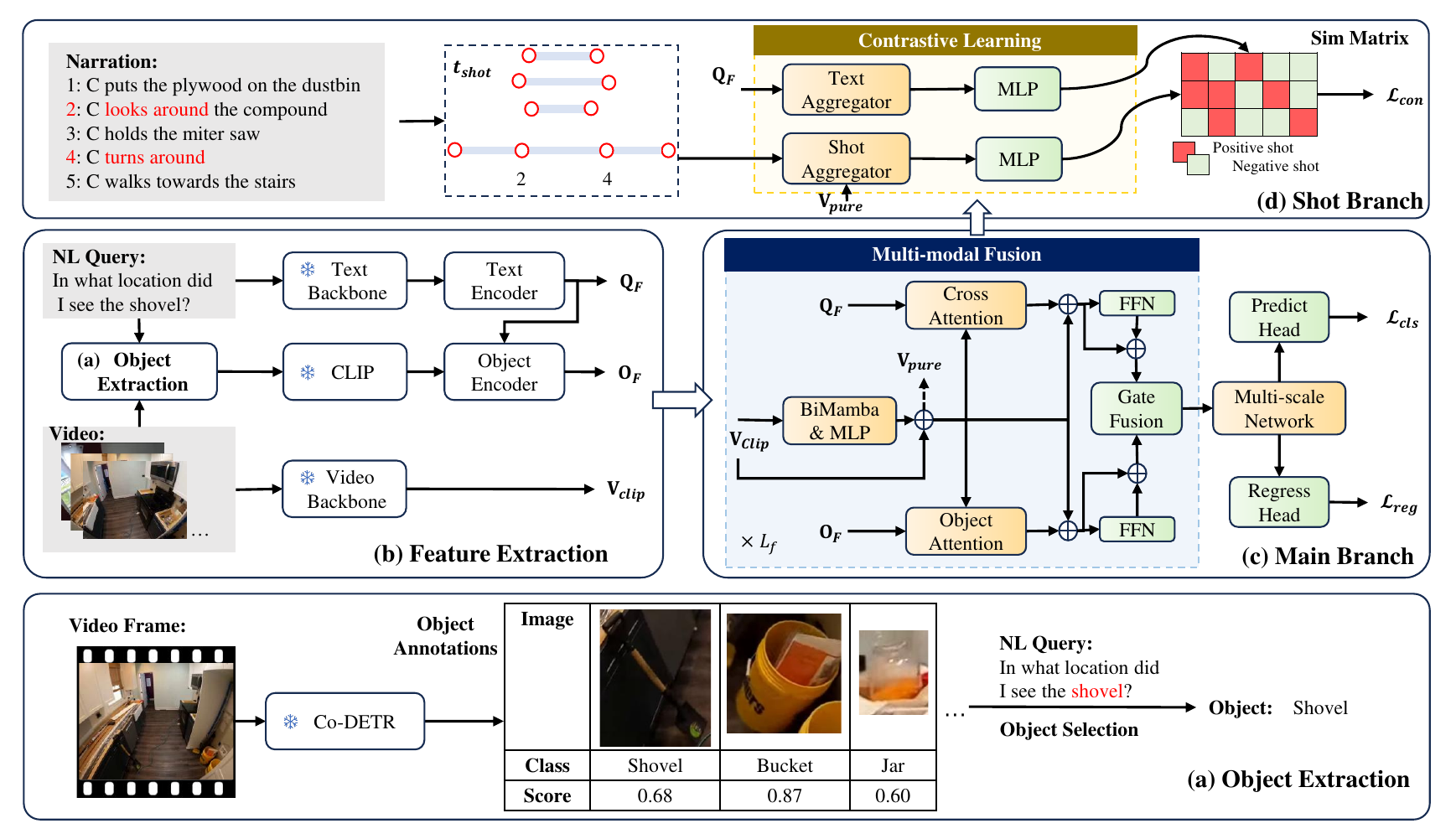}
  \vspace{-4ex}
 \caption{The framework of our OSGNet, which consists of four key components: (a) Object Extraction, which captures fine-grained object features; (b) Feature Extraction, where visual and textual cues are processed; (c) Main Branch, responsible for primary grounding tasks; and (d) Shot Branch, which leverages wearer movement dynamics and shot-level contrastive learning to improve localization accuracy.}
  \vspace{-2ex}
  \label{fig: model}
\end{figure*}
\subsection{Natural Language Query}
The NLQ task aims to accurately localize video moments that answer natural language questions in egocentric videos. 
Existing approaches primarily pursue two directions: 1) Video-Text Pretraining~\cite{kahatapitiya2024victr,zhang2024enhanced}. Due to the differences in feature distributions between egocentric and exocentric videos, several studies \cite{chen2022internvideo, lin2022egocentric, pramanick2023egovlpv2, pei2024egovideo} have fine-tuned video backbones using egocentric videos, improving the model's robustness in tasks involving egocentric video comprehension. 2) Data Enhancement. Ramakrishnan et al.~\cite{ramakrishnan2023naq} constructed a large narrative dataset of 940K video-narration pairs to pre-training models,  followed by fine-tuning on Ego4D-NLQ, yielding substantial improvements.

However, prior methods \cite{hou2023cone, hou2023groundnlq, liu2022reler} often treat NLQ as a general long-video localization problem~\cite{pan2023scanning}, overlooking the need for fine-grained object information that is crucial for NLQ. Our approach addresses this gap by extracting query-relevant fine-grained object features and integrating them into the training process, thus improving localization accuracy. Additionally, existing methods tend to overlook the frequent shot transitions inherent in egocentric videos. To mitigate this, we segment egocentric videos into distinct shots and employ contrastive learning to better align the two modalities, ultimately enhancing localization performance.



\section{Method}

\subsection{Overview}
Given a video consisting of $N$ frames, denoted as $\mathcal{V}=\{v_1,v_2,\cdots,v_N\}$, and a natural language query comprising $L$ words, denoted as $\mathcal{Q}=\{w_1,w_2,\cdots,w_L\}$, our grounding model aims to localize the specific moment within an untrimmed egocentric video that best answers the query, represented by the timestamp $[t_s,t_e]$. 

As outlined in Section~\ref{sec:intro}, existing video feature representations are often insufficient for NLQ, primarily due to the absence of fine-grained object-level information. To address this limitation, we propose a novel framework that integrates object-level details into the feature representation process, significantly enhancing the localization accuracy. 
Our approach involves extracting features from video, text, and objects, followed by a comprehensive feature encoding process, as described in Section~\ref{subsec: Feature Extraction}. 
The model architecture comprises two primary components: the main branch and the shot branch. The main branch (Section~\ref{subsec: Main Branch}) combines video, text, and object features to improve the accuracy of video localization. In contrast, the shot branch (Section~\ref{subsec: Shot Branch}) focuses exclusively on video features to generate shot representations and applies contrastive learning with text inputs, enhancing the model’s  ability to perform modality alignment. Finally, Section~\ref{subsec: Predictor and Loss} details the training and inference process of our framework.

\subsection{Feature Extraction}
\label{subsec: Feature Extraction}


\noindent \textbf{Video Feature.}
 We divide the video into non-overlapping clips, denoted as $\mathcal{C}=\{\mathcal{C}_i|\mathcal{C}_i=\{v_{(i-1)\times s+1},\cdots,v_{(i-1)\times s +s}\}\}_{i=1}^T$, where $s$ denotes the size of the sliding window and $T$ denotes the number of clips in the video. Pretrained video backbones are used to extract clip-level video features, which are then projected into a feature space, yielding the video representation as $\textbf{V}_{clip} \in \mathbb{R}^{T\times D}$, where $D$ denotes the dimensionality of the feature vectors.

\noindent \textbf{Query Feature.}
Query features are extracted using a text backbone, resulting in word-level features $\textbf{Q}_{word} \in \mathbb{R}^{L\times D_T}$, where $D_T$ is the dimensionality of the word embeddings. To capture the relationships among words in the query, we employ a multi-layer transformer, termed the text encoder, outputting the query representation $\textbf{Q}_F\in \mathbb{R}^{L\times D}$.

\noindent \textbf{Object Feature.}
To effectively capture object information, we utilize the object detector Co-DETR~\cite{zong2023detrs}, pretrained on the LVIS dataset~\cite{gupta2019lvis}, which is capable of identifying a wide range of object categories within video frames. As illustrated in Figure~\ref{fig: model}(a), we select object categories that are related to the nouns in the query and whose confidence scores exceed the threshold $\theta$.
For object representation, we encode the object categories as textual features, allowing seamless integration with the textual query.
To achieve this, we employ CLIP (ViT-B/32)~\cite{Radford2021LearningTV} to encode textual features of the detected object categories, resulting in $\textbf{O}_{clip} \in \mathbb{R}^{T\times N_o \times D_o}$,
where $N_o$ is the maximum number of objects detected in a single frame\footnote{If insufficient object representations are present, zero-padding is applied.} and $D_o$ denotes the dimensionality of the object features. 

Different objects in the same query play distinct roles in the video grounding task. For example, in the query ``How many drill bit did I remove from the drill before I moved the yellow carton?'', the beginning of the target moment should be related to the ``drill bit'' instead of the ``drill'', and the end of the target moment should before the ``carton'' appears. Therefore, we design an object encoder to refine the query-relevant object features. The object encoder is a multi-layer transformer, processing the object features in the context of the input query. To avoid confusion between object features within the same frame, we replace the conventional self-attention mechanism of transformers with a cross-attention mechanism, where object features serve as queries and query features act as keys and values. 
This architecture effectively encodes object-related information concerning the query, denoted as $\textbf{O}_F\in \mathbb{R}^{T\times N_o \times D}$.
\subsection{Main Branch}
\label{subsec: Main Branch}
Our main branch consists of three core components: a multimodal fusion module to integrate features from different modalities, a multi-scale network to generate candidate moment representations at multiple temporal scales, and task-specific heads to predict the temporal offsets and confidence scores for the localized video moments.

\noindent \textbf{Multi-modal Fusion.}
To effectively integrate fine-grained object features, we design a multi-modal fusion module that combines video, text, and object features. This module consists of multiple stacked layers, each including a bidirectional Mamba (BiMamba) block, followed by a multi-layer perceptron (MLP), cross-attention (CA) block, and Feedforward Neural Network (FFN) for query and object processing, culminating in a fusion block.

To be specific, we enhance video features using a BiMamba layer~\cite{zhuvision}, instead of traditional local self-attention, to better capture long-range dependencies within video data. The updated video features are computed as follows:
\begin{equation}
  \hat{\textbf{V}}^{(i)} =\textbf{V}^{(i)}_{f}+ MLP(BiMamba(\textbf{V}^{(i)}_{f})), 
  \label{eq:multimodelencoder1}
\end{equation}
where $\textbf{V}^{(0)}_{f}=\textbf{V}_{clip}$ and $\textbf{V}^{(i)}_{f}\in \mathbb{R}^{T \times D}$ is the output of the $i$-th layer.
Next, we apply a cross-attention block and FFN to aggregate video and query information:
\begin{equation}
\left\{
\begin{array}{l}
    \textbf{V}^{(i)}_{Q} =\hat{\textbf{V}}^{(i)}+ CA(\hat{\textbf{V}}^{(i)},\textbf{Q}_F,\textbf{Q}_F),\\[0.5em]
    \hat{\textbf{V}}^{(i)}_{Q} =\textbf{V}^{(i)}_{Q}+ FFN(\textbf{V}^{(i)}_{Q}).
    \label{eq:multimodelencoder2}
\end{array}
\right.
\end{equation}

Similarly, we aggregate video feature $\hat{\textbf{V}}^{(i)}$ and object features $\textbf{O}_F$ using a parallel cross-attention block for object features. The final output $\hat{\textbf{V}}^{(i)}_{O}$ is calculated as follows:
\begin{equation}
 \left\{
 \begin{array}{l}
     \textbf{V}^{(i)}_{O} =\hat{\textbf{V}}^{(i)}+ CA(\hat{\textbf{V}}^{(i)},\textbf{O}_F,\textbf{O}_F),\\[0.5em]
    \hat{\textbf{V}}^{(i)}_{O} =\textbf{V}^{(i)}_{O}+ FFN(\textbf{V}^{(i)}_{O}).
     \label{eq:multimodelencoder2-1}
 \end{array}
 \right.
 \end{equation}

Finally, these features are combined using a gating mechanism:
\begin{equation}
\left\{
\begin{array}{l}
    \textbf{A}=\sigma(MLP(\hat{\textbf{V}}^{(i)}_{Q} || \hat{\textbf{V}}^{(i)}_{O})),\\[0.5em]
   \textbf{V}^{(i+1)}_{f} =\textbf{A}\cdot \hat{\textbf{V}}^{(i)}_{Q}+ (\mathbf{1}
-\textbf{A})\cdot \hat{\textbf{V}}^{(i)}_{O}.
    \label{eq:multimodelencoder3}
\end{array}
\right.
\end{equation}
where $\sigma$ is sigmoid fuction, $||$ is vector concatenation.

\noindent \textbf{Multi-scale Network.}
The multi-scale network generates a feature pyramid to facilitate the grounding of video moments at various temporal scales. This network is a multi-layer transformer, with each layer including a 1D depthwise convolution before the self-attention and FFN modules to enable sequence downsampling, as described in \cite{hou2023groundnlq}. By feeding the output of the multi-modal fusion module, we can obtain the multi-scale candidate moment representations, denoted as $[\textbf{V}_{m}^{(0)},\textbf{V}_{m}^{(1)},\cdots,\textbf{V}_{m}^{(L_s)}] $,
where $\textbf{V}_{m}^{(0)}=\textbf{V}^{(L_f)}_{f}$. Here $L_f$ and $L_s$ are the number of layers in the multi-modal fusion and multi-scale network, respectively. The representation $\textbf{V}_{m}^{(j)} \in \mathbb{R}^{T/2^j \times D}$ represents the video features at the $j$-th scale, with progressively reduced sequence lengths due to downsampling at each layer. 

\noindent \textbf{Task Heads.}
The task heads are responsible for decoding the multi-scale feature pyramid into final predictions for video grounding. Specifically, we use two heads: a classification head and a regression head following previous work~\cite{mu2024snag,hou2023groundnlq,zhang2022actionformer}. Each task head is implemented using two layers of a 1D convolutional network. The classification head predicts the confidence score for each candidate moment, while the regression head predicts the offsets of the moment boundaries relative to the anchor point.

For each feature $\textbf{v}_{k}^{(j)}=\textbf{V}_{m}^{(j)}[k]\in\mathbb{R}^{D}$, $\textbf{v}_{k}^{(j)}$ represents the feature for the $k$-th anchor point in the $j$-th layer, which has $T/2^j$ anchor points with a stride of $2^j$. The corresponding timestamp is computed as $t_k^{(j)} = 2^j * k$. The classification head then predicts the confidence $c_k^{(j)}$ for $\textbf{v}_{k}^{(j)}$, and the regression head predicts normalized offsets $({s}_k^{(j)},{e}_k^{(j)})$ for $\textbf{v}_{k}^{(j)}$. The predicted video moment for this anchor point is defined by the start and end boundaries: $(t_k^{(j)} - {s}_k^{(j)}\cdot 2^j, t_k^{(j)} + {e}_k^{(j)}\cdot 2^j)$.

\noindent \textbf{Localization Loss.}
The moment localization loss, $\mathcal{L}_{ML}$, is designed to enhance model accuracy in video moment localization and is defined as follows:  
\begin{equation}
  \mathcal{L}_{ML}=\mathcal{L}_{cls}+\mathcal{L}_{reg}.
  \label{eq: moment localization}
\end{equation}
The first component  \textbf{$\mathcal{L}_{cls}$} is focal loss~\cite{ross2017focal}, which is used for classification to assess whether a proposal aligns with the query. The second component  \textbf{$\mathcal{L}_{reg}$} is Distance-IoU loss~\cite{zheng2020distance}, which refines the boundaries of the localized moment by calculating the Distance-IoU between the predicted and ground truth moment boundaries. Notably, $\mathcal{L}_{reg}$ is calculated only on positive candidates. 

\begin{table*}[t]
\caption{Performance comparison on Ego4D-NLQ.
$C$ denotes the CLIP video feature. $^*$ indicates results reproduced using the released code. $^\dagger$ refers to results that utilize the NaQ~\cite{ramakrishnan2023naq} pretraining strategy. }
  \vspace{-2ex}
  \label{tab: NLQ}
  \centering
  \resizebox{0.95\textwidth}{!}{
  \begin{tabular}{cllccccccccc}
    \toprule
    &\multirow{3}{*}{Method}&\multirow{3}{*}{Published}&\multirow{3}{*}{Feature}  & \multicolumn{4}{c}{Validation}  & \multicolumn{4}{c}{Test}        \\
    
     & &  & &\multicolumn{2}{c}{R@1}&\multicolumn{2}{c}{R@5}&\multicolumn{2}{c}{R@1}&\multicolumn{2}{c}{R@5}\\
     & &  & &0.3 & 0.5 & 0.3 & 0.5 &0.3 & 0.5 & 0.3 & 0.5 \\
    \midrule
    \multirow{8}{*}{\rotatebox{90}{Ego4D-NLQ v1}}&
    InternVideo~\cite{chen2022internvideo}& \textcolor{gray}{\small CVPRW 2022}&E+I&15.64& 10.17&24.78& 18.30&16.45 &10.06&22.95 &16.10\\
    &CONE~\cite{hou2023cone}& \textcolor{gray}{\small ACL 2023}&E&14.15&8.18&30.33&18.02&13.46&7.84&23.68&14.16\\
    &SnAG$^*$~\cite{mu2024snag}& \textcolor{gray}{\small CVPR 2024}&E&-&-&-&-&14.34&10.29&27.72&19.91\\
     &NaQ$^\dagger$~\cite{ramakrishnan2023naq}& \textcolor{gray}{\small CVPR 2023}&E+C&19.31& 11.59& 23.62 &15.75&18.46& 10.74&  21.50& 13.74\\
    &RGNet$^*$~\cite{hannan2023rgnet}& \textcolor{gray}{\small ECCV 2024}&E&16.86&10.53&34.43&21.84&16.68&10.61&27.95&17.83\\
     &RGNet$^{*\dagger}$~\cite{hannan2023rgnet}& \textcolor{gray}{\small ECCV 2024}&E&18.66&11.72&36.37&22.43&18.21&11.69&29.80&19.06\\
     &\cellcolor{gray!15}OSGNet& \cellcolor{gray!15}&\cellcolor{gray!15}E&\cellcolor{gray!15}16.13&\cellcolor{gray!15}\cellcolor{gray!15}11.28&\cellcolor{gray!15}36.78&\cellcolor{gray!15}25.63&\cellcolor{gray!15}15.23&\cellcolor{gray!15}10.71&\cellcolor{gray!15}28.27&\cellcolor{gray!15}20.32\\
     &\cellcolor{gray!15}OSGNet$^\dagger$& \cellcolor{gray!15}&\cellcolor{gray!15}E&\cellcolor{gray!15}\textbf{21.97}&\cellcolor{gray!15}\textbf{15.20}&\cellcolor{gray!15}\textbf{44.61}&\cellcolor{gray!15}\textbf{32.96}&\cellcolor{gray!15}\textbf{22.13}&\cellcolor{gray!15}\textbf{15.46}&\cellcolor{gray!15}\textbf{36.54}&\cellcolor{gray!15}\textbf{25.70}\\
    \midrule
    \multirow{7}{*}{\rotatebox{90}{Ego4D-NLQ v2}}
     &NaQ$^{*\dagger}$~\cite{ramakrishnan2023naq}& \textcolor{gray}{\small CVPR 2023}&E+I+C&24.10&15.03&29.90&20.85&21.70&13.64&25.12&16.33\\
     &ASL$^\dagger$~\cite{shao2023action}& \textcolor{gray}{\small CVPRW 2023}&E+I&22.62&15.64&46.86&32.16&24.13&15.46&34.37&23.18 \\
    & GroundNLQ$^\dagger$~\cite{hou2023groundnlq}& \textcolor{gray}{\small CVPRW 2023}&E+I  & 26.98 & 18.83 & 53.56 & 40.00 & 24.50&17.31&40.46&29.17  \\
    & EgoEnv$^\dagger$~\cite{nagarajan2024egoenv}& \textcolor{gray}{\small NeurIPS 2024}&E+I+C+EgoEnv&25.37&15.33&-&-&23.28&14.36&27.25&17.58\\
    & EgoVideo$^\dagger$~\cite{pei2024egovideo}& \textcolor{gray}{\small CVPRW 2024}&EgoVideo&28.65& 19.73& 53.30& 40.42& 25.07& 17.31& 40.88& 29.67 \\	

   & GroundVQA$^{*\dagger}$~\cite{di2024grounded}& \textcolor{gray}{\small CVPR 2024}&E+I&29.68&	20.23&	52.17&	37.83&26.67&	17.63	&	39.94	&27.70\\
     
    &\cellcolor{gray!15}OSGNet$^{\dagger}$& \cellcolor{gray!15}&\cellcolor{gray!15}E+I &\cellcolor{gray!15}\textbf{31.63} &\cellcolor{gray!15}\textbf{22.03} &\cellcolor{gray!15}\textbf{57.91} 	&\cellcolor{gray!15}\textbf{45.19} &\cellcolor{gray!15}\textbf{27.60} &\cellcolor{gray!15}\textbf{19.78} &\cellcolor{gray!15}\textbf{43.46} &\cellcolor{gray!15}\textbf{32.77}  \\
     
    \bottomrule
  \end{tabular}
  }
\end{table*}
\subsection{Shot Branch}
\label{subsec: Shot Branch}
The shot branch is specifically designed to harness the attention information embedded in egocentric videos, which reflects the wearer's focus during video capture. This is achieved by segmenting the video into semantically distinct shots, with head movement serving as a key indicator for defining these boundaries. Contrastive learning is then employed to enhance the model's ability to align cross-modal features. This shot-level understanding deepens the model's comprehension of the video structure, significantly improving video grounding performance.


\begin{figure}[t]
  \centering
  \includegraphics[width=\linewidth]{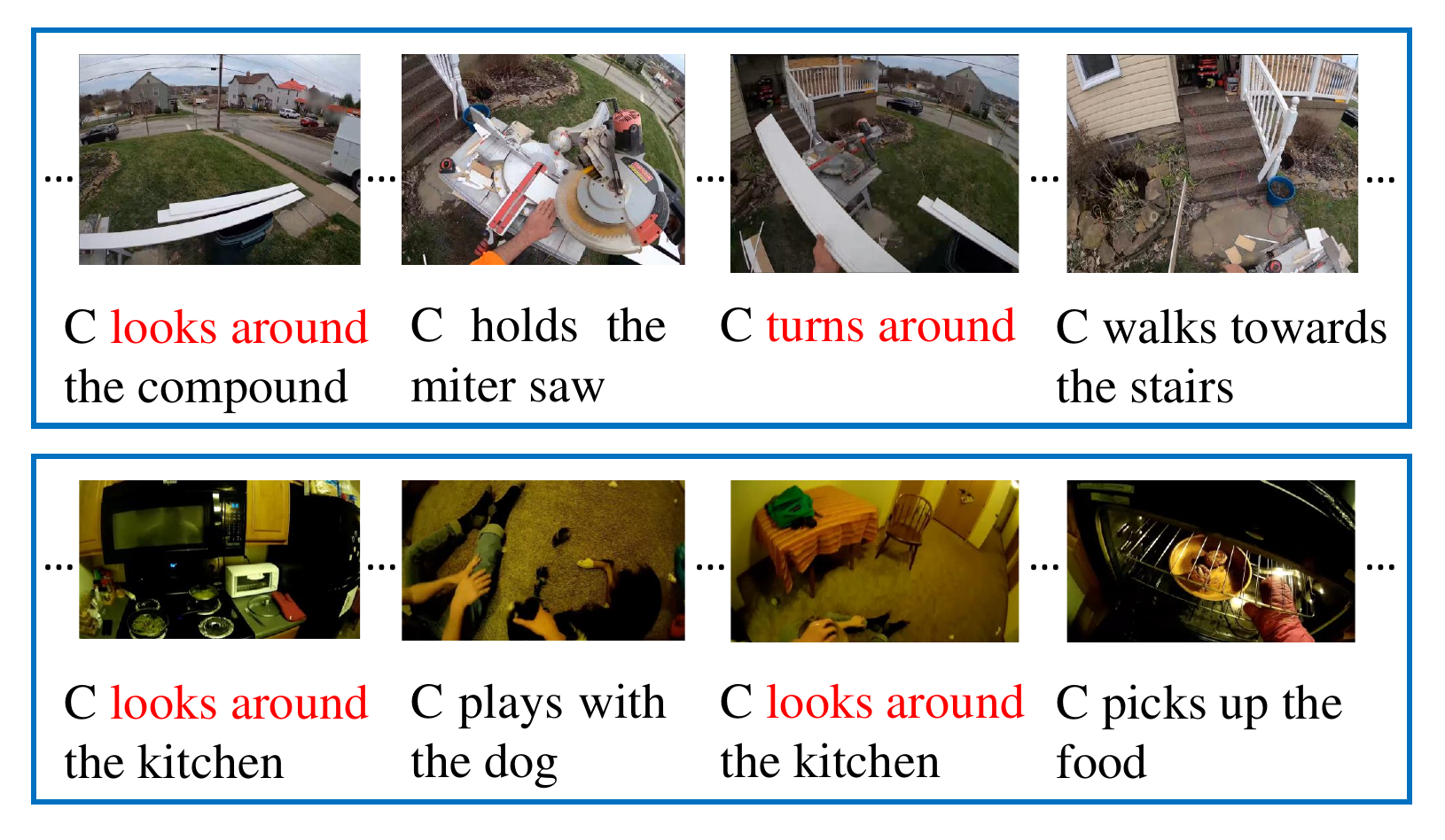}
  \vspace{-4ex}
  \caption{Illustration of captions generated by LAVILA~\cite{zhao2023learning} describing camera movements.}
  \vspace{-2ex}
  \label{fig: shot-segmentation}
\end{figure}
\noindent \textbf{Shot Segmentation.}
As illustrated in Figure~\ref{fig: shot-segmentation}, the wearer's head movements, captured by the pretrained captioning model LAVILA~\cite{zhao2023learning}, are often described in video captions with expressions such as \textit{``looks around''} or \textit{``turns around''}. These cues are used to define shot boundaries, denoted by timestamps $t_{shot} = \{(t_i, t_{i+1})\}_{i=1}^{N_S}$, where $t_1 = 0$ represents the start of the video, $t_{N_S+1}$ represents the video's end, and $\{t_i\}_{i=2}^{N_S}$ corresponds to timestamps that capture head movement activities. Here $N_S$ is the total number of shots.

\noindent \textbf{Feature Aggregation.}
To perform contrastive learning effectively, it is crucial to extract contextual video features that represent the content of each shot independently of query and object information. This is achieved by employing multiple stacked layers, each consisting of a BiMamba block integrated with an MLP. These layers share parameters with the multi-modal fusion module, enabling us to extract self-interaction features from the video, denoted as $\mathbf{V}_{pure} \in \mathbb{R}^{T \times D}$.

To aggregate shot and query features for contrastive learning, we employ a transformer encoder with an architecture similar to the Q-Former~\cite{li2023blip}. In this setup, learnable embeddings serve as queries, while the shot and query features function as keys and values.  Following this process, we apply a 1D convolution to obtain the final shot representations, denoted as $\textbf{V}_{shot}\in \mathbb{R}^{N_S \times D}$, and text representation, denoted as $\textbf{q}_{sent}\in \mathbb{R}^{D}$.


\noindent \textbf{Contrastive Learning.}
To enhance contrastive learning, we gather all shots and their associated queries within a mini-batch, denoted as $\textbf{Q}_{batch}=\{\textbf{q}_{sent}^i\}_{i=1}^m \in \mathbb{R}^{m \times D}$ and $\textbf{V}_{batch}=\{\textbf{v}_{shot}^i\}_{i=1}^n \in \mathbb{R}^{n \times D}$, respectively, where $m$ and $n$ represent the number of queries and shots in the mini-batch. Next, the text and video features are projected into a joint semantic space using separate MLPs.
These projected features are used for contrastive learning to enable the model to learn the alignment between video shots and natural language queries, thereby improving the accuracy of video moment localization.



To align shot features with textual queries, we apply contrastive learning using the InfoNCE loss~\cite{oord2018representation}, which encourages positive shot-query pairs to be closer in the embedding space while pushing apart the negative pairs. The InfoNCE loss is defined as follows:
 \begin{equation}
\begin{split}
  \mathcal{L}_{con}=\frac{1}{m}\sum_{i=1}^{m} \frac{\sum\limits_{(\textbf{q}^i_{shot},\textbf{v}^s_{shot})\in \mathcal{P}}exp(Sim(\textbf{q}^i_{sent},\textbf{v}^s_{shot})/\tau)}{\sum_{j=1}^{n} exp(Sim(\textbf{q}^i_{sent},\textbf{v}_{shot}^j)/\tau)} \\
  +\frac{1}{n}\sum_{i=1}^{n} \frac{\sum\limits_{(\textbf{q}^l_{sent},\textbf{v}^i_{shot})\in \mathcal{P}}exp(Sim(\textbf{v}^i_{shot},\textbf{q}^l_{sent})/\tau)}{\sum_{j=1}^{m} exp(Sim(\textbf{v}^i_{shot},\textbf{q}^j_{sent})/\tau)},
  \label{eq: contrastive learning5}
\end{split}
\end{equation}
where $Sim$ represents the cosine similarity, $\mathcal{P}$ is the set of positive query-shot pairs where the query's ground truth intersects with the corresponding shot, and $\tau$ is the temperature coefficient.

\subsection{Training and Inference}
\label{subsec: Predictor and Loss} 

 \noindent \textbf{Training.}
Our training is divided into pretraining and fine-tuning phases. In pretraining, we use the dataset from~\cite{ramakrishnan2023naq}, only the main branch without object cross-attention is trained with the moment localization loss.
 In fine-tuning, the final loss function combines both the moment localization loss and the contrastive learning loss, as: 
\begin{equation}
  \mathcal{L}=\frac{1}{C}(\mathcal{L}_{ML}+\lambda\mathcal{L}_{con}),
  \label{eq: contrastive learning6}
\end{equation}
where $C$ is the number of positive candidate moments (those that match the query) using momentum update, and $\lambda$ is a balancing hyperparameter.

\noindent \textbf{Inference.}
During inference, the main branch generates the start and end boundaries, along with the predicted confidence for all anchors in the pyramid. To eliminate duplicates, we apply SoftNMS~\cite{bodla2017soft} for deduplication.

\section{Experiment}
\subsection{Datasets}
Since our contributions primarily target the NLQ task, the Ego4D-NLQ dataset is our main choice. To further evaluate the versatility of our method, we also incorporated the Ego4D-Goal-Step dataset, another egocentric video grounding benchmark. Additionally, we used the TACoS dataset to enable more comprehensive comparisons.

\noindent \textbf{Ego4D-NLQ.}
Ego4D-NLQ consists of two versions. Ego4D-NLQ v1 contains 1,659 videos with 11,279, 3,874, and 4,004 video-query pairs for training, validation, and testing. 
Due to noisy data, Ego4D-NLQ v2 was released with 2,018 videos and 13,847, 4,552, and 4,004 video-query pairs for training, validation, and testing. Both versions share the same test set.

\noindent \textbf{Ego4D-Goal-Step.}
Ego4D-Goal-Step provides annotations for step grounding, which involves locating a video clip based on a step description. It contains 851 videos with 31,566, 7,696, and 5,540 video-query pairs for training, validation, and testing.

\noindent \textbf{TACoS.}
TACoS is an exocentric video moment localization dataset with 127 cooking videos. The training, validation, and test sets consist of 10,146, 4,589, and 4,083 video-query pairs. For consistency with existing works, we followed the setup in~\cite{zhang2020b}, using 9,790, 4,436, and 4,001 video-query pairs for training, validation, and testing.
\subsection{Experimental Settings}
\noindent \textbf{Implementation Details.} 
We segmented videos into clips using a sliding window of 16 frames for both window size and stride. Features are extracted using EgoVLP~\cite{lin2022egocentric} (E) and InternVideo~\cite{chen2022internvideo} (I), then concatenated as in~\cite{hou2023groundnlq}. Text features are extracted using CLIP (ViT-B/32)~\cite{Radford2021LearningTV}. For TACoS, we used C3D~\cite{Tran2015} for video features and Glove~\cite{Pennington2014} for text features to ensure a fair comparison~\cite{mu2024snag}. The object's confidence score threshold $\theta$ is 0.6. In pretraining, we used a batch size of 16, a learning rate of $8e^{-4}$, 4 warmup epochs, and 10 total epochs. In fine-tuning, warmup and total epochs are set to 4 and 10, respectively. Components that were not pretrained are initialized with weights from other pretrained structures. 
For further details, please refer to our released code.

\noindent \textbf{Baselines.} On Ego4D-NLQ v1, we compared our model with several strong baselines, including methods utilizing the NaQ pretraining strategy (NaQ~\cite{ramakrishnan2023naq} and RGNet~\cite{hannan2023rgnet}) and methods without the NaQ pretraining strategy (InternVideo~\cite{chen2022internvideo}, CONE~\cite{hou2023cone}, SnAG~\cite{mu2024snag}, and RGNet~\cite{hannan2023rgnet}), using the same validation set for a fair comparison. 
Note that RGNet~\cite{hannan2023rgnet} and SnAG~\cite{mu2024snag} employ different validation settings and have not reported performance on the Ego4D-NLQ v1 test set, we evaluated their performance in our unified setting. Further details see the \emph{Supplementary Material}.
On Ego4D-NLQ v2, we compared with strong baselines (NaQ~\cite{ramakrishnan2023naq}, ASL~\cite{shao2023action}, GroundNLQ~\cite{hou2023groundnlq}, EgoEnv~\cite{nagarajan2024egoenv}, EgoVideo~\cite{pei2024egovideo}, and GroundVQA~\cite{di2024grounded}) utilizing the NaQ pretraining strategy. As GroundVQA~\cite{di2024grounded} and NaQ~\cite{ramakrishnan2023naq} did not report performance on the Ego4D-NLQ v2 validation set, we used their published checkpoints for testing. 

On Ego4D-Goal-Step, we compared with baselines, including BayesianVSLNet~\cite{plou2024carlor}, which leverages the order prior, and EgoVideo~\cite{pei2024egovideo}, which did not use prior information.
On TACoS, we compared with the classic baselines (VSLNet~\cite{zhang2020} and 2D-TAN~\cite{zhang2020b}) and current baselines (Tri-MRF~\cite{Wang2024}, DPHANet~\cite{chen2024dphanet}, MESM~\cite{liu2024towards}, MRNet~\cite{hu2024maskable}, and SnAG~\cite{mu2024snag}).

\noindent \textbf{Evaluation Metric.}
Following previous work~\cite{hou2023cone,mu2024snag}, we utilized the metric Rank@$m$, IoU=$n$ (R@$m$, $n$), which evaluates the percentage of queries that contain at least one correct moment among the top $m$ retrieved predictions. A correct moment is defined as having an IoU greater than $n$ with the ground truth. 

\begin{table}[t]
  \caption{Performance comparison under R@1 on Ego4D-Goal-Step. $^*$ indicates results that leverage the order prior of the dataset.}
  \vspace{-2ex}
  \label{tab: Goal-Step}
  \centering
  \resizebox{0.45\textwidth}{!}{
  \begin{tabular}{lcccc}
    \toprule
    \multirow{2}{*}{Method}  & \multicolumn{2}{c}{Validation}  & \multicolumn{2}{c}{Test}        \\
    
      &0.3 & 0.5 & 0.3 & 0.5 \\
    \midrule
     EgoVideo~\cite{pei2024egovideo}&28.02&	23.66& 32.99	&25.92 \\	
    BayesianVSLNet$^*$~\cite{plou2024carlor}&18.15&8.97&35.18&20.48\\
    \rowcolor{gray!15}
    OSGNet & 29.61 & 24.94  & 32.77& 25.50  \\
    \rowcolor{gray!15}
     OSGNet$^*$ & \textbf{42.61} & \textbf{35.38}  & \textbf{38.83}& \textbf{30.16} \\
    \bottomrule
  \end{tabular}
  }
\end{table}
%
\begin{table}[t]
  \caption{Performance comparison on TACoS. $^*$ indicates results that utilize the E+I video feature.}
  \vspace{-2ex}
  \label{tab: TACoS}
  \centering
  \resizebox{0.45\textwidth}{!}{
  \begin{tabular}{llcccc}
    \toprule
    \multirow{2}{*}{Method}&\multirow{2}{*}{Published} &\multicolumn{2}{c}{R@1}&\multicolumn{2}{c}{R@5}\\
    & &0.3 & 0.5 & 0.3 & 0.5   \\
    \midrule
    VSLNet~\cite{zhang2020}& \textcolor{gray}{\small ACL 2020}&29.61&24.27&-&-\\
     2D-TAN~\cite{zhang2020b}& \textcolor{gray}{\small AAAI 2020}&37.29&25.32&57.81&45.04\\
     Tri-MRF~\cite{Wang2024}& \textcolor{gray}{\small TMM 2024}&52.44&41.49&76.01&63.46\\
     DPHANet~\cite{chen2024dphanet}& \textcolor{gray}{\small TMM 2024}&47.01&34.12&-&-\\
     MESM~\cite{liu2024towards}& \textcolor{gray}{\small AAAI 2024}&52.69&39.52&-&-\\
     MRNet~\cite{hu2024maskable}& \textcolor{gray}{\small ACMMM 2024}&55.41&38.54&77.18&64.78\\
     SnAG~\cite{mu2024snag}& \textcolor{gray}{\small CVPR 2024}&56.44&	44.86& 81.15	&70.66 \\	
     \rowcolor{gray!15}
     OSGNet& & \textbf{57.57} & \textbf{48.18} & \textbf{82.02} &	\textbf{72.05} \\
    \rowcolor{gray!15}
     OSGNet$^*$& &66.43 & 55.77 & 87.16 &	79.45 \\
     
    \bottomrule
  \end{tabular}
  }
\end{table}
\subsection{Performance Comparison}

\noindent \textbf{On Ego4D-NLQ.}
Table~\ref{tab: NLQ} presents the performance comparison on the validation and the test sets of Ego4D-NLQ. On Ego4D-NLQ v2, our model outperforms all compared methods across all metrics. 
Specifically, compared to the strong baseline GroundVQA, our model achieves an absolute improvement of over 1.5\% in R@1, 0.5 on both the validation and test sets.
 On Ego4D-NLQ v1, our model surpasses the baseline RGNet~\cite{hannan2023rgnet} in most metrics, including the challenging R@1, 0.5.  Our carefully designed architecture results in a nearly 4.0\% absolute improvement in R@1, 0.5 after pretraining. In contrast, RGNet~\cite{hannan2023rgnet} shows considerably less benefit from pretraining, likely due to its dual-branch fusion structure. 
 
\noindent \textbf{On Ego4D-Goal-Step.}
Table~\ref{tab: Goal-Step} presents the performance of our model compared to existing methods on the validation and test sets of Ego4D-Goal-Step. Our OSGNet outperforms all competing methods across all metrics on the validation set. Without any prior information, our model achieves an absolute improvement of approximately 1.59\% in R@1, 0.3 on the validation set, compared to the strong baseline EgoVideo. When incorporating the order prior of the steps, as done by BayesianVSLNet~\cite{plou2024carlor}, our model further improves by around 3.65\% in R@1, 0.3 on the test set, surpassing BayesianVSLNet~\cite{plou2024carlor}.

\noindent \textbf{On TACoS.}
Table~\ref{tab: TACoS} presents the performance comparison of our model with existing methods on the test set of TACoS. Our OSGNet outperforms all competing methods across all metrics on the test set when using the C3D feature. Compared to the strong baseline SnAG, our model achieves an absolute improvement of approximately 3.32\% in the challenging metric R@1, 0.5. Additionally, when concatenating the InternVideo~\cite{chen2022internvideo} and EgoVLP~\cite{lin2022egocentric} video features, along with the pretraining strategy, our model achieves 66.43\% IoU=0.3 and 55.77\% IoU=0.5 at R@1.


\noindent \textbf{Summary.} Comparing results across different datasets, we observed that our model performs lowest on Ego4D-NLQ, followed by Ego4D-Goal-Step, and highest on TACoS under the same settings. We attributed the relatively lower performance on egocentric video grounding datasets, compared to exocentric ones, to the inherent complexity of understanding egocentric video content. Furthermore, NLQ presents additional challenges due to its emphasis on grounding fine-grained objects within the background. Despite this, our model excels in R@1, 0.5 on both Ego4D-NLQ and TACoS, demonstrating the effectiveness of our proposed approach.

     

\begin{figure}[t]
  \centering
  \includegraphics[width=\linewidth]{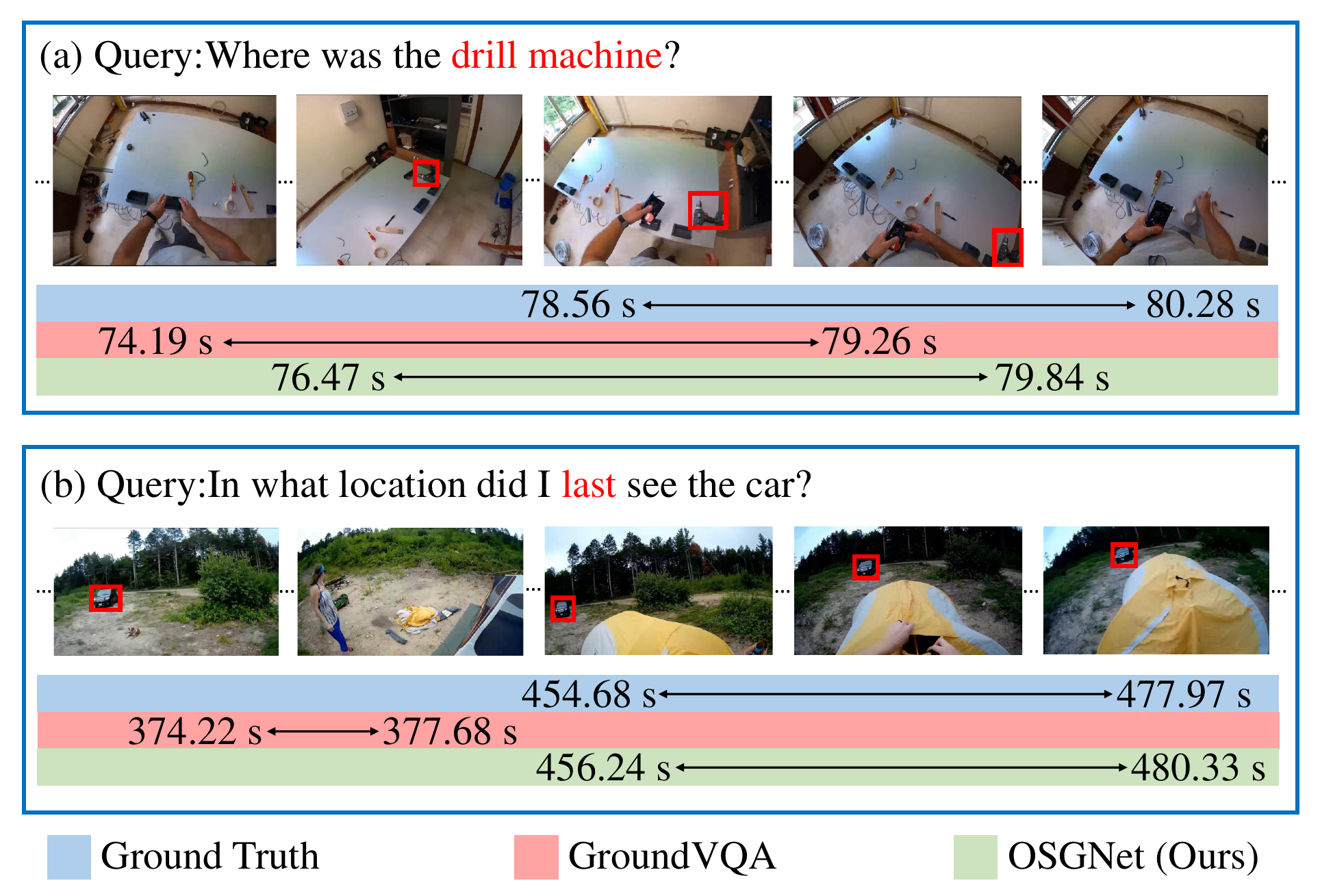}
  \vspace{-4ex}
  \caption{Qualitative comparison with GroundVQA on Ego4D-NLQ.}
  \vspace{-2ex}
  \label{fig: visualize}
\end{figure}

\subsection{Ablation Study}
\noindent \textbf{On Model Structure.} We conducted ablation studies on the validation set of Ego4D-NLQ v2 to assess the contributions of key components in our model. Specifically, we examined the impact of removing fine-grained object features by excluding the object branch and fusion block from the multi-modal encoder, as well as the effect of removing shot-level contrastive learning by eliminating the $\mathcal{L}_{con}$ term from the total loss function $\mathcal{L}$. Table~\ref{tab: component} presents the results of these ablations. Removing $\mathcal{L}_{con}$ leads to a performance drop of around 0.4\% across all R@1 metrics, indicating that the shot branch plays a crucial role in enhancing the model's ability to learn better video representations. Additionally, removing the object feature results in over 2.0\% decline in the R@1 metric, highlighting the importance of object-level information for the NLQ task.


\noindent \textbf{On Shot Segmentation.} Table~\ref{tab: shot slicing method} displays the effect of our shot segmentation method on the validation set of Ego4D-NLQ. We hypothesized that camera movement, in addition to head rotation, may be relevant for shot segmentation. To test this, we selected movement-related verbs from the generated captions that occur more than 100 times, including ``walks'', ``moves around'', ``rides'', ``runs'', ``cycles'', ``jogs'', and ``jumps''. 
Based on these frequencies, we divided the verbs into three groups: head rotation verbs (R: ``turns around'', ``looks around''), high-frequency movement verbs (HM: ``walks'', ``moves around''), and other main movement verbs (RM: ``rides'', ``runs'', ``cycles'', ``jogs'', ``jumps''). We first examined segmentation using only R and HM and observed that the model using R for segmentation typically performs better on R@1, which aligns with the main verbs in the captions. When combining R with HM and RM, the results are comparable to using R alone, leading us to adopt R for segmentation due to its efficiency. 


\noindent \textbf{On Object Features.} Table~\ref{tab: object features} illustrates the impact of different types of object features on the validation set of Ego4D-NLQ. The results indicate that text-based object features outperform image-based object features by 0.65\% in R@1, 0.3. We suppose that text-based object features align more effectively with the video modality.

    
    
\subsection{Qualitative Analysis}
We compared our model with the strong baseline GroundVQA \cite{di2024grounded} on the Ego4D-NLQ validation set through qualitative analysis. As shown in Figure~\ref{fig: visualize}(a), OSGNet accurately grounds the video moment where the ``drill machine'' is located. In contrast, GroundVQA struggles with fine-grained object localization due to the limitations of its video features, resulting in less accurate predictions.
In Figure~\ref{fig: visualize}(b), OSGNet accurately selects the moment when the car was last seen, demonstrating stronger semantic alignment. On the other hand, while GroundVQA can identify the car's appearance in the video, it fails to predict the correct moment due to its limited video understanding.


\begin{table}[t]
  \caption{Ablation studies on model structure.}
  \vspace{-2ex}
  \label{tab: component}
  \centering
  \begin{tabular}{cccccc}
    \toprule
    \multirow{2}{*}{$\mathcal{L}_{con}$}&\multirow{2}{*}{Object} &\multicolumn{2}{c}{R@1}&\multicolumn{2}{c}{R@5}\\
    &&0.3 & 0.5 & 0.3 & 0.5  \\
     \midrule
     \ding{55}&\ding{55} &28.87&	19.60& 54.53	&41.78 \\
    \ding{51}&\ding{55} &29.22&19.99& 55.12	&41.78 \\
    \ding{55}&\ding{51} &31.26&	21.64& \textbf{58.19}	&\textbf{45.19} \\	
\rowcolor{gray!15}
     \ding{51}&\ding{51}&  \textbf{31.63} & \textbf{22.03} & 57.91 &	\textbf{45.19} \\
     
    \bottomrule
  \end{tabular}
\end{table}
\begin{table}[t]
  \caption{Ablation studies on verb selection for shot segmentation.}
  \vspace{-2ex}
  \label{tab: shot slicing method}
  \centering
  \begin{tabular}{ccccccc}
    \toprule
    \multirow{2}{*}{R}&\multirow{2}{*}{HM}&\multirow{2}{*}{RM} &\multicolumn{2}{c}{R@1}&\multicolumn{2}{c}{R@5}\\
    & & &0.3 & 0.5 & 0.3 & 0.5  \\
     \midrule
     \ding{51}&\ding{51}&\ding{51} &31.61&	21.99& \textbf{58.52}	&44.86 \\	
     \ding{51}&\ding{51}&\ding{55} &31.22&	21.84& 56.83	&44.44 \\	
    \ding{55}&\ding{51}&\ding{55} &30.91&21.20&58.28&\textbf{45.41}\\
    \rowcolor{gray!15}
     \ding{51}&\ding{55}&\ding{55}&  \textbf{31.63} & \textbf{22.03} & 57.91 &	45.19 \\
     
    \bottomrule
  \end{tabular}
\end{table}

\begin{table}[t]
  \caption{Ablation studies on the object features.}
  \vspace{-2ex}
  \label{tab: object features}
  \centering
  \begin{tabular}{ccccc}
    \toprule
    \multirow{2}{*}{Object Features} &\multicolumn{2}{c}{R@1}&\multicolumn{2}{c}{R@5}\\
     &0.3 & 0.5 & 0.3 & 0.5  \\
     \midrule
     Image &30.98&	21.68	&	\textbf{58.22}&	45.14\\	
    \rowcolor{gray!15}
     Text&  \textbf{31.63} & \textbf{22.03} & 57.91 &	\textbf{45.19} \\
    \bottomrule
  \end{tabular}
\end{table}
\section{Conclusion}
In this paper, we propose OSGNet 
for egocentric video grounding. 
Our key improvements are twofold: 1) capturing object information to improve video representation, better aligning it with text queries; and 2) capturing shot information for contrastive learning, implicitly enhancing video representation. Through comprehensive experiments, we validate the effectiveness of these improvements. 
\newline

\noindent \textbf{Acknowledgements.} This work is partially supported by the National Natural Science Foundation of China, No.62236003, No.62476071, No.62376140, and No.U23A20315; the Shenzhen College Stability Support Plan, No.GXWD20220817144428005; the Science and Technology Innovation Program for Distinguished Young Scholars of Shandong Province Higher Education Institutions, No.2023KJ128, and the Special Fund for Taishan Scholar Project of Shandong Province.

{
    \small
    \bibliographystyle{ieeenat_fullname}
    \bibliography{main}
}
\clearpage
\setcounter{page}{1}
\setcounter{section}{0}
\renewcommand{\thesection}{\Alph{section}}
\maketitlesupplementary
In the supplementary material, we first outlined the baseline settings on Ego4D-NLQ to establish their validity(Section~\ref{sec: Reproduction Result Settings}). 
We then conducted ablation experiments from scratch to eliminate potential bias from pretraining, and also conducted additional ablation experiments to verify the rationality of our design(Section~\ref{sec: Ablation Study}). 
Next, we categorized Ego4D-NLQ by question templates and compared model performance across these categories to the strong baseline, GroundVQA~\cite{di2024grounded}, highlighting our improvement in background object-related query localization (Section~\ref{sec: templates}). To further demonstrate the effectiveness of our model, we provided comprehensive visualizations illustrating the diverse data and our model's predictions (Section~\ref{sec: Visualization}). 
Finally, we introduced the model structure and other implementation details (Section~\ref{sec: Implementation Details}).

\section{Baseline Settings}
\label{sec: Reproduction Result Settings}
\subsection{On Ego4D-NLQ v1}
In Ego4D-NLQ v1, there is a significant amount of noisy data with ground truth durations of 0, resulting in predicted outputs consistently yielding an IoU of 0, making accurate localization impossible. As a result, RGNet~\cite{hannan2023rgnet} and SnAG~\cite{mu2024snag} remove these noisy samples from the validation set. However, for a fair comparison across all methods, we evaluated all models on the original validation set, including the noisy samples.

\paragraph{RGNet.}
Since RGNet~\cite{hannan2023rgnet} did not release any checkpoints trained from scratch, we retrained the model (No.2 in Table~\ref{tab: RGNet-scratch}). For the pretraining setting, we utilized the fine-tuned checkpoints published by RGNet~\cite{hannan2023rgnet} for testing on the original validation sets (No.2 in Table~\ref{tab: RGNet}). RGNet~\cite{hannan2023rgnet} removes $N^R_{noisy}=341$ noisy samples with ground truth durations of 0, along with $N^R_{add}=4$ additional samples. Assuming that the predictions on these $N^R_{add}$ samples are correct, we can adjust the evaluation result as follows:
\begin{equation}
\left\{
\begin{array}{l}
    N^{R}=N_{val}-N^R_{noisy}-N^R_{add},\\[0.5em]
    m^R_{cor}=\frac{m^R_{ori}*N^{R}+N^R_{add}}{N_{val}},
\end{array}
\right.
\end{equation}
where $N^R$ is the number of validation samples used in RGNet, $N_{val}=3874$ is the number of total samples in the Ego4d-NLQ v1 validation set, $m^R_{cor}$ represents the evaluation result after adjustment (No.3 in Table~\ref{tab: RGNet-scratch} and Table~\ref{tab: RGNet}), and $m^R_{ori}$ is the result reported in the origin paper  (No.1 in Table~\ref{tab: RGNet-scratch} and Table~\ref{tab: RGNet}).
\begin{table}[h]
 \caption{The results of RGNet trained from scratch on the validation set under different settings.}
\vspace{-2ex}
  \label{tab: RGNet-scratch}
  \centering
  \begin{tabular}{cccccc}
    \toprule
    \multirow{2}{*}{No.} &\multirow{2}{*}{Setting} &\multicolumn{2}{c}{R@1}&\multicolumn{2}{c}{R@5}\\
     & &  0.3 & 0.5 & 0.3 & 0.5  \\
    \midrule
    1&Original&18.28&12.04&34.02&22.89\\  
    2&Reproduce&16.86&10.53&34.43&21.84\\
    3&Correction&16.76&11.07&31.09&20.95\\
    \bottomrule
  \end{tabular}
\end{table}
\begin{table}[h]
  \caption{The result of RGNet with NaQ pretraining strategy on the validation set under different settings.}
\vspace{-2ex}
  \label{tab: RGNet}
  \centering
  \begin{tabular}{cccccc}
    \toprule
    \multirow{2}{*}{No.}  &\multirow{2}{*}{Setting}&\multicolumn{2}{c}{R@1}&\multicolumn{2}{c}{R@5}\\
     &  & 0.3 & 0.5 & 0.3 & 0.5  \\
    \midrule
   1& Original&20.63&12.47&41.67&25.08\\
   2& Checkpoint&18.66&11.72&36.37&22.43\\
    3 & Correction&18.90&11.46&38.06&22.95\\
    \bottomrule
  \end{tabular}
\end{table}

\paragraph{SnAG.}
Since SnAG~\cite{mu2024snag} has not released complete annotations for the validation set, we used the published checkpoints to evaluate the test set (SnAG in Table 1 in the \emph{Manuscript}) and applied a formula to adjust the results for the validation set (No.2 in Table~\ref{tab: SnAG}). SnAG~\cite{mu2024snag} removes $N^S_{noisy}=341$ noisy samples with ground truth durations of 0, along with $N^S_{add}=35$ additional samples. Assuming the predictions on these $N^S_{add}$ samples are correct, we adjusted the results as follows:
\begin{equation}
\left\{
\begin{array}{l}
    N^{S}=N_{val}-N^S_{noisy}-N^S_{add},\\[0.5em]
    m^S_{cor}=\frac{m^S_{ori}*N^{S}+N^S_{add}}{N_{val}},
\end{array}
\right.
\end{equation}
where $N^S$ represents the number of validation samples used in SnAG, $m^S_{cor}$ are the results after adjustment (No.2 in Table~\ref{tab: SnAG}), and $m^S_{ori}$ are the results reported in the origin paper (No.1 in Table~\ref{tab: SnAG}).

\begin{table}[t]
  \caption{The result of SnAG~\cite{mu2024snag} on the validation set under different settings.}
\vspace{-2ex}
  \label{tab: SnAG}
  \centering
  \begin{tabular}{cccccc}
    \toprule
    \multirow{2}{*}{No.} &\multirow{2}{*}{Setting} &\multicolumn{2}{c}{R@1}&\multicolumn{2}{c}{R@5}\\
     & &  0.3 & 0.5 & 0.3 & 0.5  \\
    \midrule
   1 &Original&15.87&11.26&38.26&27.16\\
    2&Correction&15.23&11.07&35.45&25.43\\
    \bottomrule
  \end{tabular}
\end{table}

\section{Ablation Study}
\label{sec: Ablation Study}
\subsection{On Scratch}

\begin{table}[t]
  \caption{Ablation studies on model structure for Ego4D-NLQ v1.}
  \label{tab: scratch-NLQ v1}
  \vspace{-2ex}
  \centering
  \begin{tabular}{cccccc}
    \toprule
    \multirow{2}{*}{$\mathcal{L}_{con}$}&\multirow{2}{*}{Object} &\multicolumn{2}{c}{R@1}&\multicolumn{2}{c}{R@5}\\
    &&0.3 & 0.5 & 0.3 & 0.5  \\
     \midrule
     \ding{55}&\ding{55} &9.53&	6.61& 27.21	&18.61 \\
    \ding{51}&\ding{55} &13.37&8.98& 32.21	&22.02 \\
    \ding{55}&\ding{51} &14.07	&10.12	&33.84	&24.08 \\	
\rowcolor{gray!15}
     \ding{51}&\ding{51}&  \textbf{16.13}&\textbf{11.28}&\textbf{36.78}&\textbf{25.63} \\
     
    \bottomrule
  \end{tabular}
\end{table}
\begin{table}[t]
  \caption{Ablation studies on model structure for Ego4D-NLQ v2.}
  \label{tab: scratch-NLQ v2}
  \vspace{-2ex}
  \centering
  \begin{tabular}{cccccc}
    \toprule
    \multirow{2}{*}{$\mathcal{L}_{con}$}&\multirow{2}{*}{Object} &\multicolumn{2}{c}{R@1}&\multicolumn{2}{c}{R@5}\\
    &&0.3 & 0.5 & 0.3 & 0.5  \\
     \midrule
     \ding{55}&\ding{55} &13.27&	9.16& 36.78	&25.92 \\
    \ding{51}&\ding{55} &17.29&11.86& 40.55	&28.80 \\
    \ding{55}&\ding{51} &17.40&	12.10	&	41.10&	29.92\\
\rowcolor{gray!15}
     \ding{51}&\ding{51}&  \textbf{18.74}&\textbf{12.72}&\textbf{41.92}&\textbf{30.34} \\
     
    \bottomrule
  \end{tabular}
\end{table}
\begin{figure*}[t]
  \centering
  \includegraphics[width=0.85\textwidth]{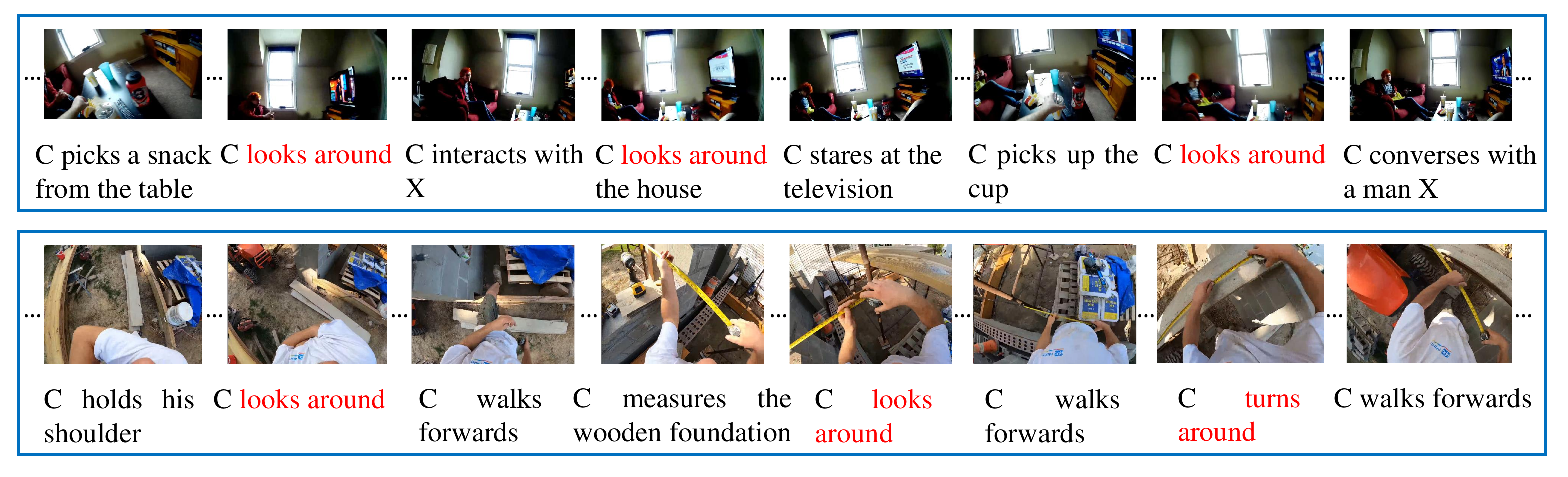}
  \vspace{-2ex}
  \caption{Illustration of captions generated by LAVILA~\cite{zhao2023learning} describing camera movements.}
  \label{fig: long-shot-segmentation}
\end{figure*}
\begin{figure*}[t]
  \centering
  \includegraphics[width=0.7\textwidth]{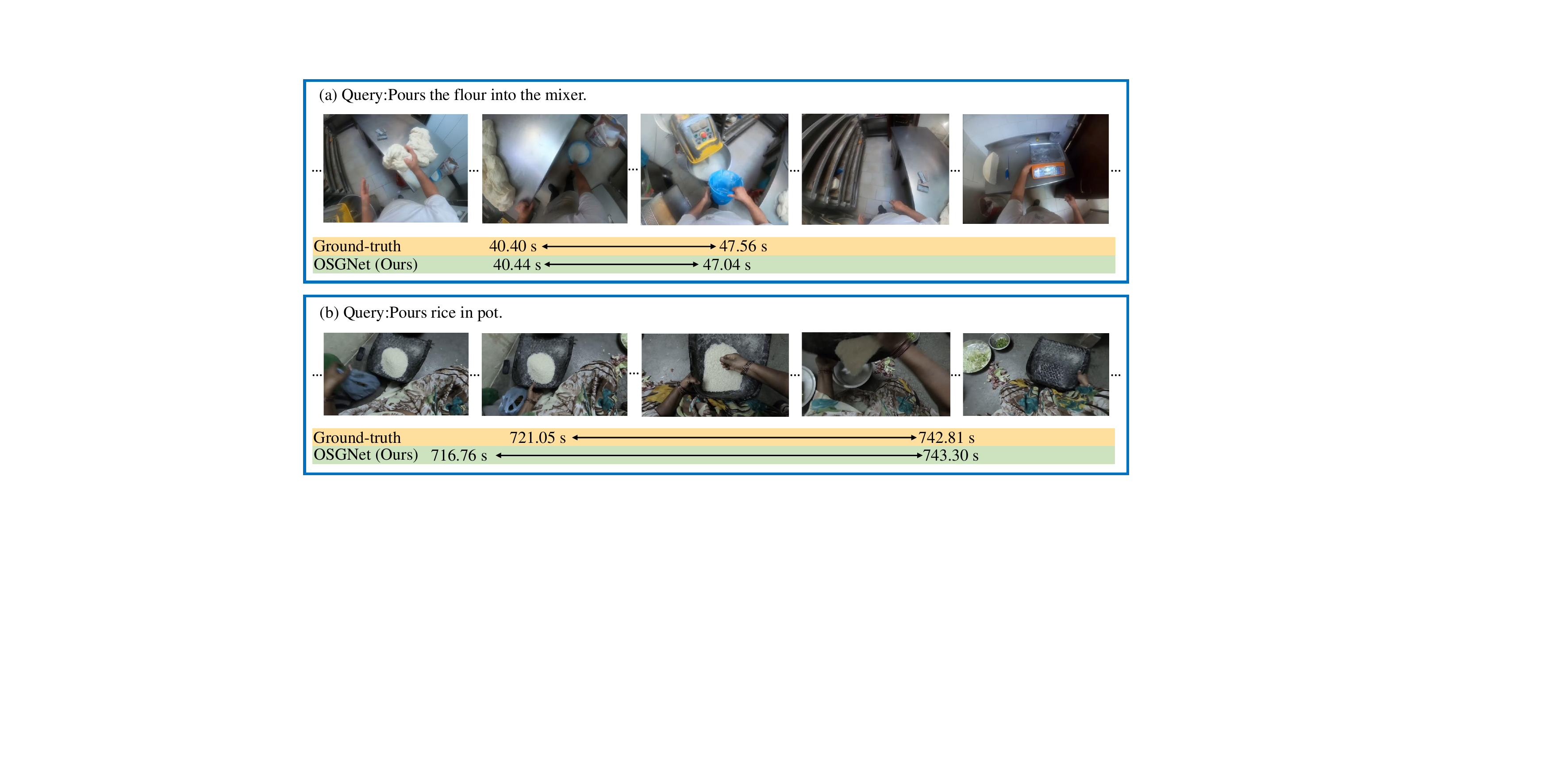}
  \vspace{-2ex}
  \caption{Illustration of grounding results on Ego4D-Goal-Step.}

  \label{fig: goalstep-vis}
\end{figure*}

To further demonstrate the efficacy of our module, we conducted ablation experiments on the Ego4D-NLQ dataset while excluding the NaQ pretraining strategy (NaQ~\cite{ramakrishnan2023naq}). The results for Ego4D-NLQ v1 and Ego4D-NLQ v2 are presented in Table~\ref{tab: scratch-NLQ v1} and Table~\ref{tab: scratch-NLQ v2}, respectively.

In Table~\ref{tab: scratch-NLQ v1}, removing $\mathcal{L}_{con}$ results in a 1.16\% performance drop at R@1, 0.5, while excluding the object feature leads to a 2.30\% decrease. Similarly, Table~\ref{tab: scratch-NLQ v2} shows that ablating the object feature and the shot branch causes declines of 0.86\% and 0.62\% in R@1, 0.5, respectively. These results underscore the critical role of the shot-level branch in enhancing video representation learning and highlight the importance of object-level information for the NLQ task.

Compared to the ablation experiments in the \emph{Manuscript}, the shot branch exhibits a more significant improvement when pretraining is excluded. We attribute this to two factors. First, during pretraining, our model undergoes extensive semantic alignment training, causing the enhancements provided by the shot branch to overlap with those already acquired, thereby yielding limited additional gains. Second, because the shot branch is not included in the pretraining phase, its data alignment remains misaligned with that of the main branch, further constraining its performance improvements. 
\begin{table}[t]
  \caption{Ablation studies on shot segmentation.}
  \vspace{-2ex}
  \label{tab: Ablation studies on shot segmentation}
  \centering
  \begin{tabular}{ccccc}
    \toprule
    \multirow{2}{*}{Method} &\multicolumn{2}{c}{R@1}&\multicolumn{2}{c}{R@5}\\
     &0.3 & 0.5 & 0.3 & 0.5  \\
     \midrule
     Average &31.22&	21.42	&	\textbf{58.22}&	44.95\\	
     
    \rowcolor{gray!15}
     LAVILA&  \textbf{31.63} & \textbf{22.03} & 57.91 &	\textbf{45.19} \\
    \bottomrule
  \end{tabular}
\end{table}
\subsection{On Shot Segmentation}

In Table 5 of the \emph{Manuscript}, we validate our design by analyzing high-frequency movement-related verbs. Additionally, Table~\ref{tab: Ablation studies on shot segmentation} presents result from segmenting videos with an average shot length of 13 seconds without using verbs. In this experiment, R@1, 0.5 reaches 21.42\%, and R@1, 0.3 reaches 31.22\%, with an average drop of only 0.5 points. These results confirm that the performance improvement is not driven by any bias from movement-related verbs.
\subsection{On Main Branch}
\begin{table}[t]
  \caption{Ablation studies on the self-mixer in the main branch.}
  \vspace{-2ex}
  \label{tab: Ablation studies on the self-mixer in the main branch}
  \centering
  \begin{tabular}{ccccc}
    \toprule
    \multirow{2}{*}{Self-mixer} &\multicolumn{2}{c}{R@1}&\multicolumn{2}{c}{R@5}\\
     &0.3 & 0.5 & 0.3 & 0.5  \\
     \midrule
     Self-attention &28.14&	19.66	&	55.38&	42.64\\	
     
    \rowcolor{gray!15}
     BiMamba&  \textbf{31.63} & \textbf{22.03} & \textbf{57.91} &	\textbf{45.19} \\
    \bottomrule
  \end{tabular}
\end{table}
Table~\ref{tab: Ablation studies on the self-mixer in the main branch} shows the impact of our self-mixer on the Ego4D-NLQ validation set. We opted Mamba over self-attention to enhance long-term temporal modeling, which results in a 2.37\% improvement in R@1, 0.5. 
\begin{table}[t]
  \caption{Ablation studies on the multi-scale network in the main branch.}
  \vspace{-2ex}
  \label{tab: Ablation studies on the multi-scale network in the main branch}
  \centering
  \begin{tabular}{ccccc}
    \toprule
    \multirow{2}{*}{$L_s$} &\multicolumn{2}{c}{R@1}&\multicolumn{2}{c}{R@5}\\
     &0.3 & 0.5 & 0.3 & 0.5  \\
     \midrule
     0 &28.30&	18.85	&	53.54&	40.29\\	
     1&29.90&	20.21	&	55.58	&42.18\\
     2&30.60&	20.47	&	57.12&	43.41\\
    4&	31.24&	21.44&	57.49&	43.96\\

    \rowcolor{gray!15}
     6&  \textbf{31.63} & \textbf{22.03} & 57.91 &	\textbf{45.19} \\
     8&	31.20	&21.59	&	\textbf{58.06}&	44.75\\
    \bottomrule
  \end{tabular}
\end{table}
Additionally, Table~\ref{tab: Ablation studies on the multi-scale network in the main branch} shows that a six-layer multi-scale network provides the best performance.

\begin{table*}[t]
  \caption{Performance comparison on different question templates in Ego4D-NLQ v2.}
  \label{tab: templates}
  \vspace{-2ex}
  \centering
  \begin{tabular}{cp{3cm}ccccccccc}
    \toprule
    \multirow{3}{*}{\parbox{2cm}{\centering Category}}&\multirow{3}{*}{Template}&\multirow{3}{*}{No.} &\multicolumn{4}{c}{OSGNet(Ours)}&\multicolumn{4}{c}{GroundVQA}\\ 
    &&&\multicolumn{2}{c}{R@1}&\multicolumn{2}{c}{R@5}&\multicolumn{2}{c}{R@1}&\multicolumn{2}{c}{R@5}\\
    &&&0.3 & 0.5 & 0.3 & 0.5&0.3 & 0.5 & 0.3 & 0.5  \\
     \midrule
     \multirow{9}{*}{\parbox{2cm}{\centering Interactive Objects}}&\hangindent=1.2em \hangafter=1 \raggedright 
  \cellcolor{gray!15} 1. Where is object X before / after event Y?&\cellcolor{gray!15} \multirow{3}{*}{750}&\cellcolor{gray!15} \multirow{3}{*}{37.07}&\cellcolor{gray!15} \multirow{3}{*}{23.33}&\cellcolor{gray!15} \multirow{3}{*}{64.53}&\cellcolor{gray!15} \multirow{3}{*}{48.93}&\cellcolor{gray!15}  \multirow{3}{*}{40.67}   &\cellcolor{gray!15}   \multirow{3}{*}{25.33}   &\cellcolor{gray!15}   \multirow{3}{*}{63.73}   &\cellcolor{gray!15}   \multirow{3}{*}{44.93}\\
     
     &\hangindent=1.2em \hangafter=1 \raggedright 2. What did I put in X?& \multirow{2}{*}{546}&\multirow{2}{*}{33.52}&\multirow{2}{*}{24.73}&\multirow{2}{*}{65.75}&\multirow{2}{*}{54.03}&\multirow{2}{*}{34.62}   &  \multirow{2}{*}{25.64}   &  \multirow{2}{*}{64.10}   &  \multirow{2}{*}{49.63} \\
     
     &\hangindent=1.2em \hangafter=1 \raggedright 
     \cellcolor{gray!15}3. What X did I Y?&\cellcolor{gray!15}350&\cellcolor{gray!15}39.14&\cellcolor{gray!15}28.86&\cellcolor{gray!15}66.86&\cellcolor{gray!15}54.00&\cellcolor{gray!15}38.57   &\cellcolor{gray!15}  28.29   &\cellcolor{gray!15}  67.43   &\cellcolor{gray!15}  50.29\\
     
     &\hangindent=1.2em \hangafter=1 \raggedright 4. What X is Y?&332&26.81&17.77&48.19&37.65& 26.20   &  17.17   &  42.47   &  28.31 \\
     
     &\hangindent=1.2em \hangafter=1 \raggedright 
     \cellcolor{gray!15}5. State of an object&\cellcolor{gray!15}235&\cellcolor{gray!15}42.55&\cellcolor{gray!15}30.21&\cellcolor{gray!15}70.21&\cellcolor{gray!15}56.17&\cellcolor{gray!15}39.15   &\cellcolor{gray!15}  26.38   &\cellcolor{gray!15}  60.43   &\cellcolor{gray!15}  43.83 \\
     
     &\hangindent=1.2em \hangafter=1 \raggedright 6. Where did I put X?&\multirow{2}{*}{725}&\multirow{2}{*}{32.83}&\multirow{2}{*}{19.86}&\multirow{2}{*}{58.48}&\multirow{2}{*}{43.31}& \multirow{2}{*}{31.31}   &  \multirow{2}{*}{18.48}   &  \multirow{2}{*}{54.90}   &  \multirow{2}{*}{36.69} \\
     
     \midrule
     
     \multirow{7}{*}{\parbox{2cm}{\centering Background Objects}}&\hangindent=1.2em \hangafter=1 \raggedright 
     \cellcolor{gray!15}7. Where is object X?&\cellcolor{gray!15}\multirow{2}{*}{552}&\cellcolor{gray!15}\multirow{2}{*}{18.30}&\cellcolor{gray!15}\multirow{2}{*}{14.49}&\cellcolor{gray!15}\multirow{2}{*}{42.75}&\cellcolor{gray!15}\multirow{2}{*}{30.98}&\cellcolor{gray!15}\multirow{2}{*}{11.59}   &\cellcolor{gray!15}   \multirow{2}{*}{7.97}   &\cellcolor{gray!15}  \multirow{2}{*}{29.35}   &\cellcolor{gray!15}  \multirow{2}{*}{18.66}\\
     
     &\hangindent=1.2em \hangafter=1 \raggedright 8. How many X’s?&386&40.41&31.35&64.25&56.48& 33.16   &  27.20   &  52.33   &  44.56\\
     
     &\hangindent=1.2em \hangafter=1 \raggedright 
     \cellcolor{gray!15}9. In what location did I see object X?&\cellcolor{gray!15}\multirow{3}{*}{359}&\cellcolor{gray!15}\multirow{3}{*}{20.33}&\cellcolor{gray!15}\multirow{3}{*}{15.60}&\cellcolor{gray!15}\multirow{3}{*}{43.45}&\cellcolor{gray!15}\multirow{3}{*}{33.70}&\cellcolor{gray!15}\multirow{3}{*}{11.70}   &\cellcolor{gray!15}   \multirow{3}{*}{7.52}   & \cellcolor{gray!15} \multirow{3}{*}{31.75}   & \cellcolor{gray!15} \multirow{3}{*}{23.12} \\
     
     &\hangindent=1.2em \hangafter=1 \raggedright 10. Where is my object X?&\multirow{2}{*}{72}&\multirow{2}{*}{16.67}&\multirow{2}{*}{15.28}&\multirow{2}{*}{40.28}&\multirow{2}{*}{33.33}&\multirow{2}{*}{18.06}   &   \multirow{2}{*}{13.89}   &  \multirow{2}{*}{33.33}   &  \multirow{2}{*}{23.61} \\
     
     \midrule
     
     \multirow{8}{*}{\parbox{2cm}{\centering People}} &\hangindent=1.2em \hangafter=1 \raggedright \cellcolor{gray!15} 11. Who did I interact with when I did activity X?&\cellcolor{gray!15}\multirow{4}{*}{115}&\cellcolor{gray!15}\multirow{4}{*}{31.30}&\cellcolor{gray!15}\multirow{4}{*}{20.00}&\cellcolor{gray!15}\multirow{4}{*}{53.91}&\cellcolor{gray!15}\multirow{4}{*}{39.13}&\cellcolor{gray!15} \multirow{4}{*}{26.96}   &\cellcolor{gray!15}  \multirow{4}{*}{15.65}   &\cellcolor{gray!15}  \multirow{4}{*}{53.04}   &\cellcolor{gray!15}  \multirow{4}{*}{33.91} \\
     
     &\hangindent=1.2em \hangafter=1 \raggedright 12. Who did I talk to in location X?&\multirow{2}{*}{91}&\multirow{2}{*}{28.57}&\multirow{2}{*}{20.88}&\multirow{2}{*}{54.95}&\multirow{2}{*}{46.15}& \multirow{2}{*}{30.77}   &  \multirow{2}{*}{23.08}   &  \multirow{2}{*}{57.14}   &  \multirow{2}{*}{41.76}  \\
     
     &\hangindent=1.2em \hangafter=1 \raggedright \cellcolor{gray!15} 13. When did I talk to or interact with person with role X?&\cellcolor{gray!15}\multirow{4}{*}{22}&\cellcolor{gray!15}\multirow{4}{*}{22.73}&\cellcolor{gray!15}\multirow{4}{*}{13.64}&\cellcolor{gray!15}\multirow{4}{*}{54.55}&\cellcolor{gray!15}\multirow{4}{*}{36.36}&\cellcolor{gray!15}\multirow{4}{*}{18.18}   &\cellcolor{gray!15}  \multirow{4}{*}{13.64}   &\cellcolor{gray!15}  \multirow{4}{*}{50.00}   &\cellcolor{gray!15}  \multirow{4}{*}{22.73}\\
     
     \midrule
     
     \parbox{2cm}{\centering None}&\hangindent=1.2em \hangafter=1 \raggedright 14. None&17&29.41&23.53&58.82&35.29&35.29   &  35.29   &  47.06   &  47.06\\
     
     \midrule
     
     \parbox{2cm}{\centering Total}&&4552&31.63 & 22.03 & 57.91 &	45.19&29.68&	20.23&	52.17&	37.83\\
    \bottomrule
  \end{tabular}
\end{table*}

\section{On Question Template}
\label{sec: templates}
We conducted an in-depth analysis of our model’s performance across various question templates in Ego4D-NLQ v2, as summarized in Table~\ref{tab: templates}. The question templates are divided into three categories:  queries about interacted objects, queries about background objects, and queries focusing on interactions involving people without specific objects. Queries with missing template information are classified as ``None'' and are relatively few. 

From the table, we could find that the model's performance on questions involving background objects is significantly lower compared to the other two categories, highlighting the difficulty of understanding background elements in Ego4D-NLQ v2. Moreover, when comparing our model with GroundVQA~\cite{di2024grounded}, we observed an improvement of 1.39-8.08\%  in R@1, 0.5 for the background object categories, emphasizing that our enhancements significantly improve background object understanding.

\begin{figure*}
    \centering
    \begin{subfigure}{0.8\linewidth}
        \centering
        \includegraphics[width=\textwidth]{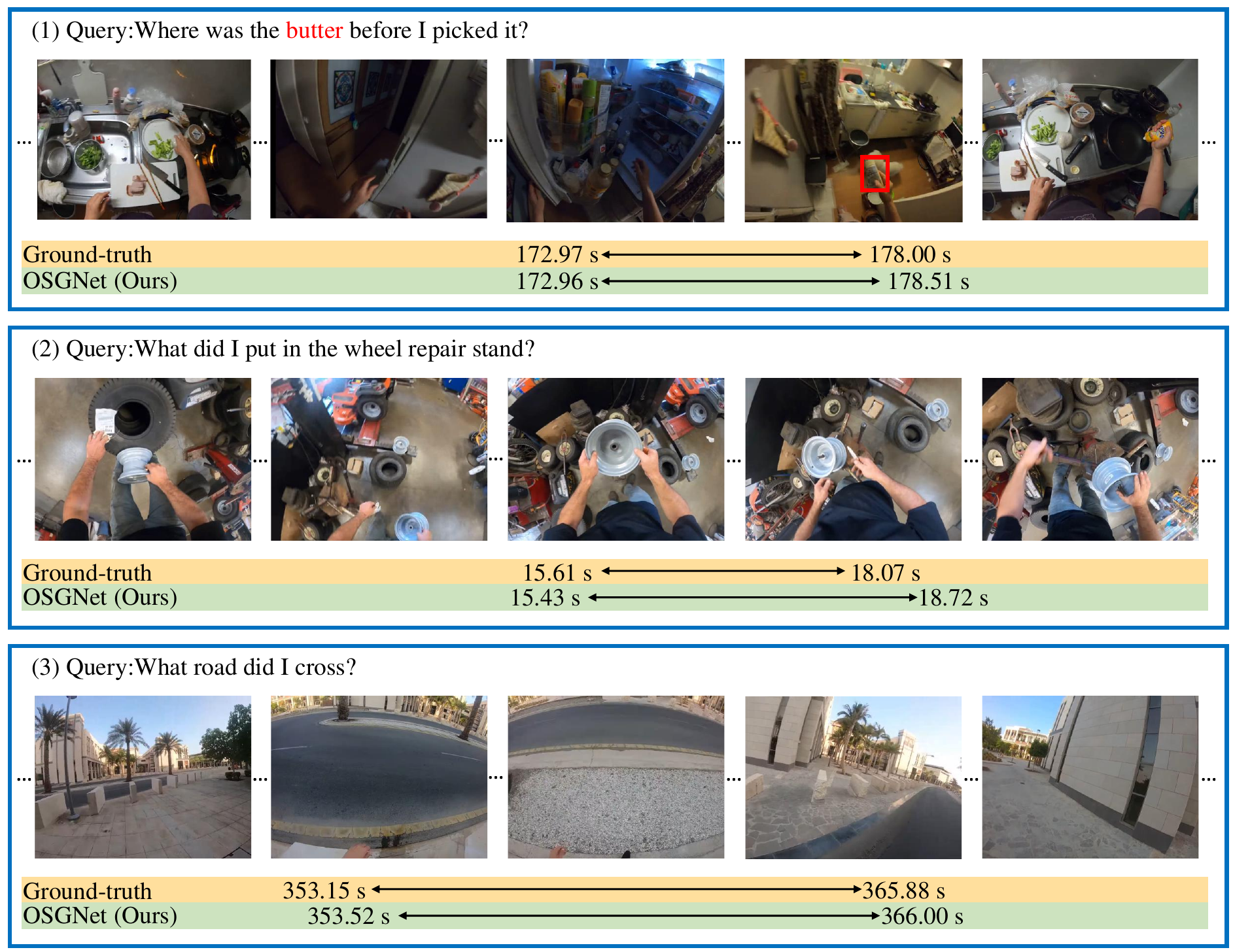} 

    \end{subfigure}
    
    \begin{subfigure}{0.8\linewidth}
        \centering
        \includegraphics[width=\textwidth]{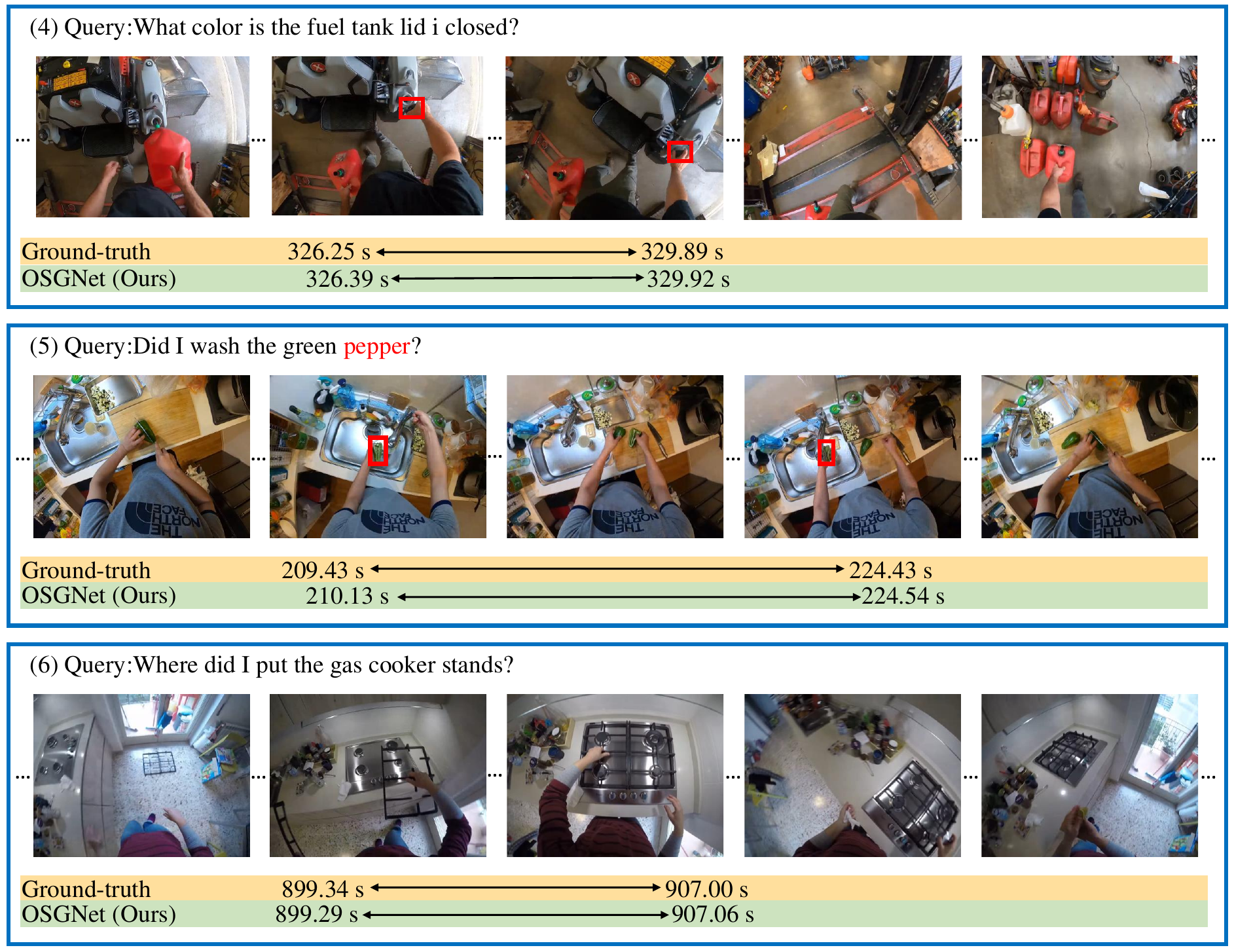} 
        
    \end{subfigure}
   \caption{Illustration of grounding results on the question templates 1-6 in Ego4D-NLQ.}

    \label{fig: templates1-6}
\end{figure*}
\begin{figure*}
    \centering
    \begin{subfigure}{0.8\linewidth}
        \centering
        \includegraphics[width=\textwidth]{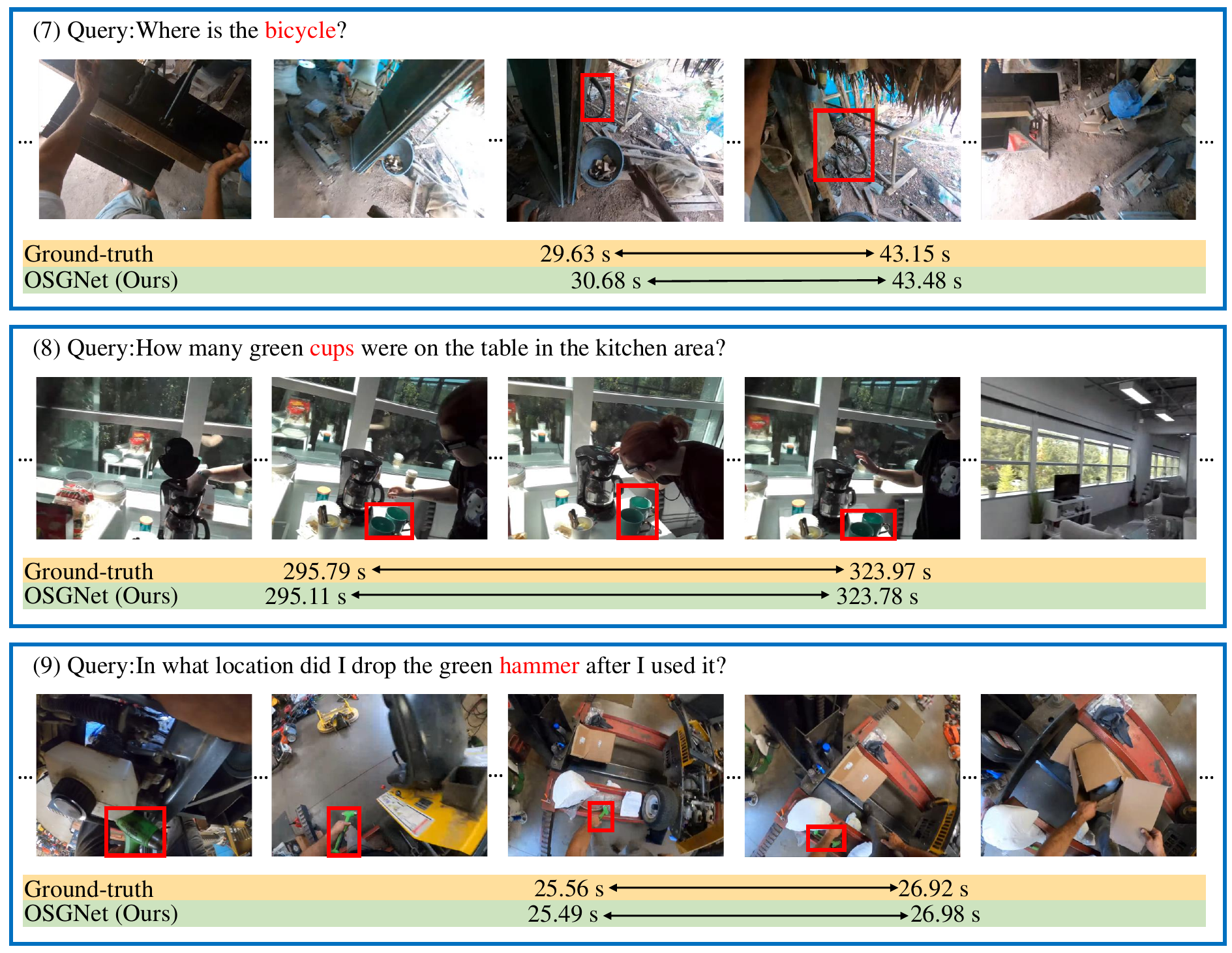} 

    \end{subfigure}
    
    \begin{subfigure}{0.8\linewidth}
        \centering
        \includegraphics[width=\textwidth]{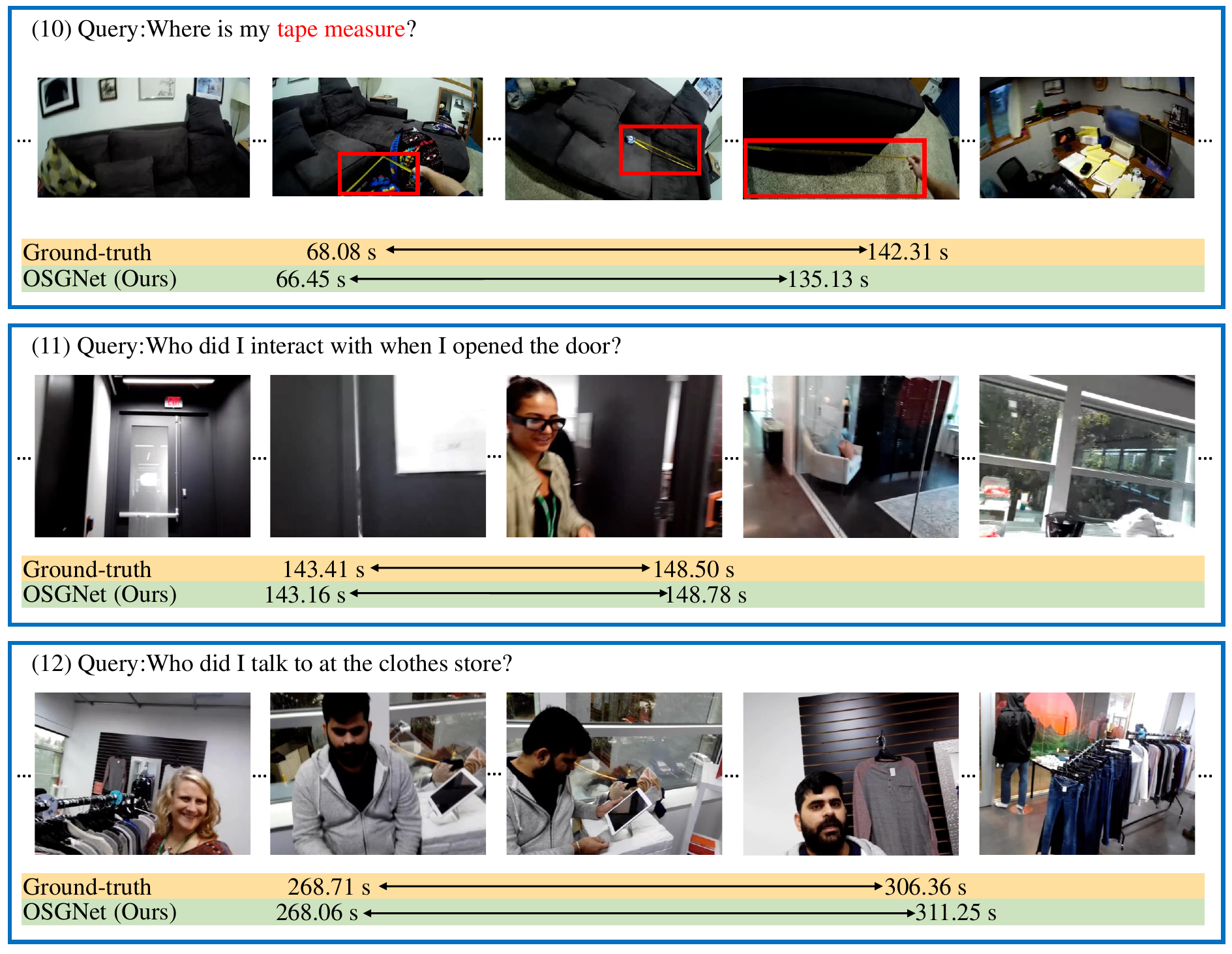} 
        
    \end{subfigure}
   \caption{Illustration of grounding results on the question templates 7-12 in Ego4D-NLQ.}

    \label{fig: templates7-12}
\end{figure*}
\begin{figure*}[t]
  \centering
  \includegraphics[width=0.8\linewidth]{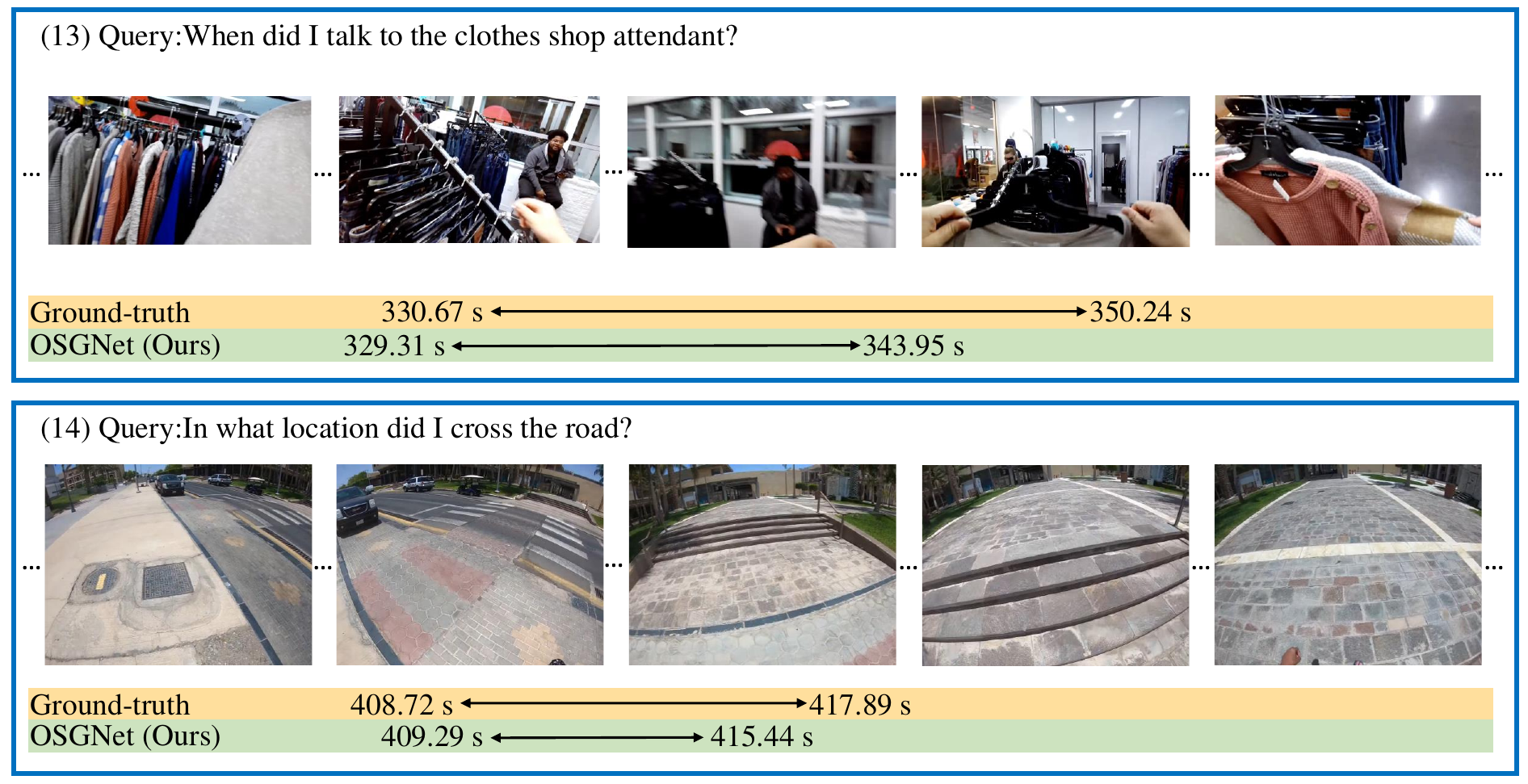}

  \caption{Illustration of grounding results on the question templates 13-14 in Ego4D-NLQ.}

  \label{fig: templates13-14}
\end{figure*}
\section{Qualitative Experiments}
\label{sec: Visualization}

\paragraph{Shot Visualization.}
We demonstrated the effect of shot segmentation in longer videos. As shown in Figure~\ref{fig: long-shot-segmentation}, camera movement is prevalent in egocentric videos, with the average shot length, using our shot-slicing strategy, being 13 seconds.
 
\paragraph{Ego4D-Goal-Step.}
 
As shown in Figure~\ref{fig: goalstep-vis}, our model accurately locates events in Ego4D-Goal-Step, demonstrating its strong ability to localize actions and objects.

\paragraph{Ego4D-NLQ.}
As shown in Figures~\ref{fig: templates1-6}, \ref{fig: templates7-12}, and \ref{fig: templates13-14}, our model accurately locates moments corresponding to various types of questions, demonstrating its robust and versatile localization capabilities.

\section{Implementation Details}
\label{sec: Implementation Details}
\subsection{Model Structure} 
The text encoder consists of 4 transformer layers, the same number as the object encoder. The multi-modal fusion module contains $L_f = 4$ layers, while the multi-scale network has $L_s = 6$ layers. Additionally, the aggregators in the shot branch each use a single-layer network.
\subsection{Object Detection}
Popular object detectors like SAM and Grounding DINO were tested but struggled with detecting fine-grained objects. Therefore, Co-DETR was chosen, an open-source model that performs exceptionally well on LVIS, a dataset with over 1,000 object categories. To match objects in the query, we use spaCy to extract nouns and measure their semantic similarity to object classes. Of the 22,396 queries in the Ego4D-NLQ v2 dataset we used, 22,313 were found to contain nouns by spaCy, and 18,871 matched the object categories in LVIS.

\subsection{Symbol $t_{shot}$ of Figure 2}
The blue line at the bottom represents the entire video timeline. The numbers 2 and 4, positioned below the red circles, indicate the segmentation timestamps corresponding to the 2nd and 4th narrations, which include movement-related verbs. Three lines above the blue line represent the three segmented shots. 

\subsection{Computational Efficiency }
Pretraining takes 4 L20 GPUs for 3 days, while fine-tuning on Ego4D-NLQ requires 2 L20 GPUs for 3 hours. The model has 122M trainable parameters, but for inference, this reduces to 106M due to the shot branch being used only during training. On the NLQ v2 validation set, with an average video length of 9 minutes, inference speed is 19.45 video-text pairs per second using 1 L20 GPU. Text feature extraction with CLIP is very fast, processing thousands of sentences per second. Feature extraction times for LAVILA, InternVideo, and Co-DETR are 1/7, 1/3, and 1.5 times the video duration, respectively, using 1 L20 GPU.


\end{document}